\documentclass[conference]{IEEEtran}
\IEEEoverridecommandlockouts

\usepackage{cite}
\usepackage{amsmath,amssymb,amsfonts}
\usepackage{algorithmic}
\usepackage{graphicx}
\usepackage{enumitem}
\usepackage{textcomp}
\usepackage{xcolor}
\usepackage{multirow}
\usepackage{makecell}
\usepackage{fancyhdr}
\def\BibTeX{{\rm B\kern-.05em{\sc i\kern-.025em b}\kern-.08em
    T\kern-.1667em\lower.7ex\hbox{E}\kern-.125emX}}
\usepackage{xspace}
\usepackage{caption3}

\newcommand{\textsub}[1]{$_{\text{#1}}$}
\newcommand{\classifier}{PICROT\xspace}

\title{Sinkhorn Divergence of Topological Signature Estimates for Time Series Classification}
\author{
	\IEEEauthorblockN{Colin Stephen}
	\IEEEauthorblockA{
		School of Computing, Electronics and Mathematics,
		Coventry University UK \\
		Email: colin.stephen@coventry.ac.uk
	}
}

\begin{document}
\maketitle

\thispagestyle{fancy}
\lhead{Accepted Version}
\rhead{2018 17th IEEE International Conference on Machine Learning and Applications}
\lfoot{\copyright 2018 IEEE\\ https://doi.org/10.1109/ICMLA.2018.00113}

\begin{abstract}
Distinguishing between classes of time series sampled from dynamic systems is a common challenge in systems and control engineering, for example in the context of health monitoring, fault detection, and quality control. The challenge is increased when no underlying model of a system is known, measurement noise is present, and long signals need to be interpreted. In this paper we address these issues with a new non parametric classifier based on topological signatures. Our model learns classes as weighted kernel density estimates (KDEs) over persistent homology diagrams and predicts new trajectory labels using Sinkhorn divergences on the space of diagram KDEs to quantify proximity. We show that this approach accurately discriminates between states of chaotic systems that are close in parameter space, and its performance is robust to noise.
\end{abstract}

\IEEEpeerreviewmaketitle

\section{Introduction}


Automatic labelling of time series is a significant challenge in many scientific applications: predicting cardiac pathologies from electrocardiogram (ECG) traces, the imminent failure of engine components from vibration measurements, or the type of a star from its light spectrum variations for example. Features that are invariant under transformations such as nonlinear warping in the time domain are often important for effective predictions in these contexts because measurement sampling rates may differ between samples, the same physical processes may evolve at different rates for different samples, or the morphology of the signal during important events may be the deciding factor between classes. So a wide range of methods for time series classification based on different invariances are available, offering a variety of performance characteristics suited to different applications \cite{Bagnall2017,Esling2012}. In the case of dynamic systems, it is also known that signal decomposition and interpretation methods such as spectral and cepstral analysis, and phase space reconstructions using Takens embedding theorem, provide useful features to interpret and compare the states of systems \cite{Kantz2004,Koopmans1995,Randall2017}.

The approach taken in this paper compares global descriptors of time series, like symbolic aggregate approximations (SAX) or bags of patterns (BOP) do, but without the need to decompose signals and using a metric that is invariant to nonlinear time domain transformations, like dynamic time warping (DTW) is. In particular \emph{we show how to summarise the topological features of classes of time series in a concise way, and how to quickly quantify similarity between the topology of an unlabelled time series and these class summaries.}

Our method applies techniques from topological data analysis (TDA) in the form of persistent homology (PH) to characterize signals \cite{Carlsson2009,Edelsbrunner2014}. The central object of interest in PH is the persistence diagram (PD) or barcode which provides a concise and stable formal representation of the topological features present in a data set at \emph{all metric scales} simultaneously \cite{Cohen-Steiner2010,Ghrist2008a}. This global multi-scale perspective allows TDA to expose features otherwise overlooked by conventional nonlinear dimensionality reduction techniques. For an introduction see \cite{Ghrist2008a}, for algorithmic aspects \cite{Edelsbrunner2010}, and for examples of metric space representations on the space of PDs needed to use TDA in off-the-shelf machine learning pipelines see \cite{Adams2017,Reininghaus2015}.


\section{Related Work}

Time series have previously been analysed using TDA, largely via PH of point clouds constructed using delay embeddings \cite{Khasawneh2016,Perea2014,Pereira2015a,Seversky2016}. However delay embeddings require heuristic estimation of the embedding dimension and delay size to be computed beforehand, increase the influence of noise, and lead to a rapid increase in the complexity of computing PH of filtrations on the point cloud, requiring subsampling techniques such as the witness complex \cite{Alexander2015,DeSilva2004,Sanderson2017}. One previous study avoids embeddings but relies on a coarse grained statistic derived from PDs to characterise time series, the persistent entropy \cite{Rucco2017}. Our method is similar to this since we use a filtration on the time series directly, however we use a metric on the space of persistence diagrams directly rather than on the space of persistent entropy histograms.

Our method uses persistence images (PIs) as its underlying stable representation of PH, however rather than using these as feature vectors in a support vector classifier as in \cite{Adams2017}, we apply a distance metric to them. Following \cite{Chazal2018} we treat PIs as kernel estimators of the density of expected PDs for a class, and to measure distance between these density estimates and new PIs we use the Sinkhorn divergence \cite{Cuturi2013a}. The latter is an upper bound approximation to the Wasserstein distance between distributions which can be computed very quickly \cite{Altschuler2017,Blondel2017,Solomon2015}. The Sinkhorn divergence has recently been applied to scalable clustering and averaging of large PDs \cite{Lacombe2018}, but not yet to classification using topological features.

Much previous work has been done on characterising and comparing trajectories of dynamical systems, but this often focuses on methods for distinguishing between chaotic and non-chaotic (periodic, quasi-periodic, intermittent) behaviour, for example by analysing spectra of Lyapunov exponents or textures of recurrence plots \cite{Marwan2007,Wolf1985}. The method we develop here is suited to this type of classification but to showcase its fine-grained capabilities we focus numerical experiments on showing that it can distinguish between different chaotic regimes that lie very close to one another in parameter space.

\section{Persistent Homology and Optimal Transport}
\label{sec:background}

We first outline persistent homology in terms of sub level sets of functions, then the persistence image representation of persistence diagrams, before introducing entropy regularized optimal transport metrics between probability distributions. These topics form the backbone of the classifier pipeline defined and applied in following sections.

\subsection{Persistent Homology and Persistence Images}

\begin{figure}[htb]
	\caption{Mapping critical points of a function (left) to (birth, death) pairs in a persistence diagram (right). From Edelsbrunner and Harer, ``Persistent Homology -- A Survey'' \cite{Edelsbrunner2008}.}
	\label{fig:sublevel_pd}
	\includegraphics[width=0.48\textwidth]{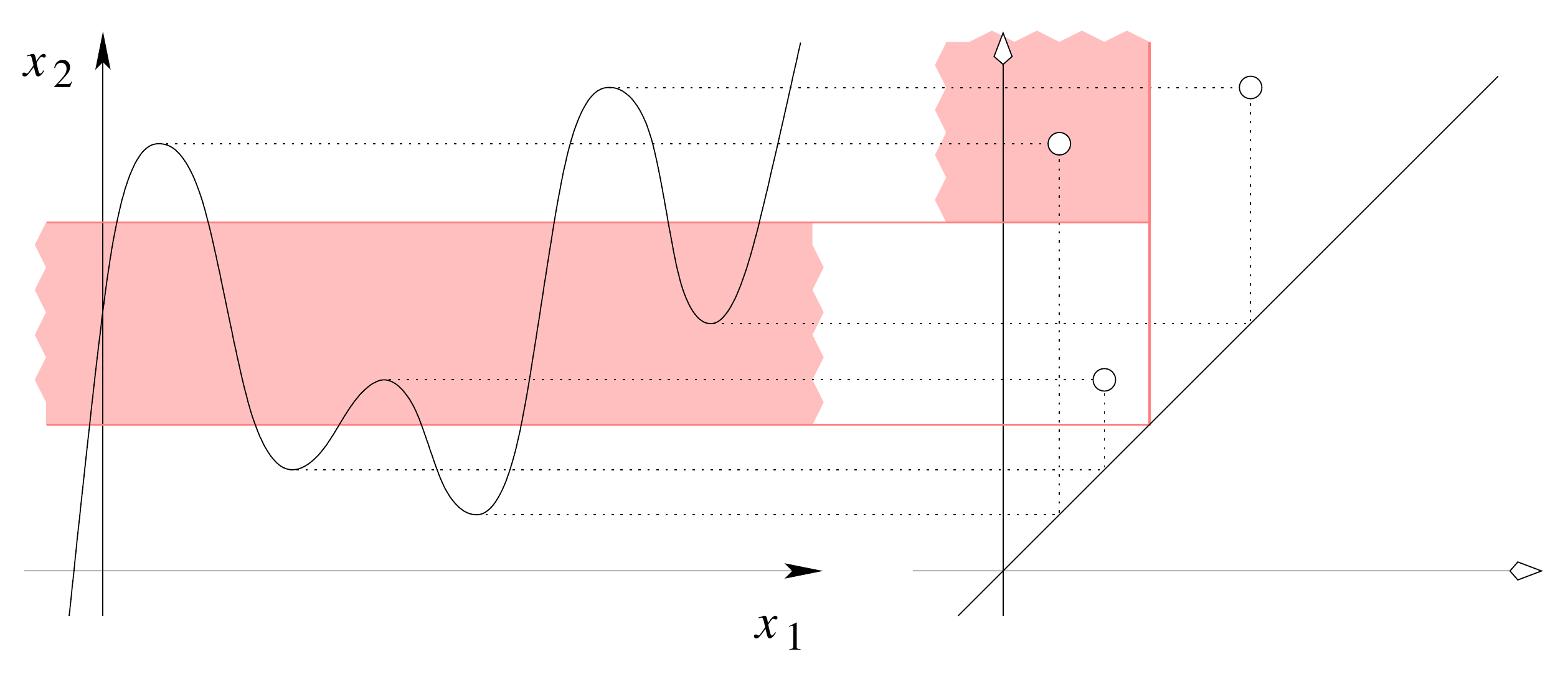}
\end{figure}

Given a bounded continuous function $f:\mathbb{X}\to\mathbb{R}$ on a topological space $\mathbb{X}$ define sublevel sets $\mathbb{X}_a:=f^{-1}(-\infty,a]$ for each $a$ in $\mathbb{R}$. Then given $a\leq b$ the inclusion $\mathbb{X}_a \subseteq \mathbb{X}_b$ induces a homomorphism of homology groups: $\mathbf{f}_l^{a,b}:H_l(\mathbb{X}_a)\to H_l(\mathbb{X}_b)$ for each dimension $l$. Under mild conditions on $f$, for any $\delta>0$ the homomorphism $\mathbf{f}_l^{c-\delta,c}$ is \emph{not} an isomorphism for only finitely many values of $c\in\mathbb{R}$ for all $l$, and $H_l(\mathbb{X}_a)$ is finitely generated \cite{Cohen-Steiner2007}. This guarantees that the following procedure results in a finite data structure: step through the values of $c$ at which the homology of $\mathbb{X}_a$ changes and record the maximal intervals $[b,d]\subset\mathbb{R}$ such that homology classes appearing in some $\mathbb{X}_a$ live in precisely one interval and no where else. The filtration values $b,d\in\mathbb{R}$ describing each such interval are often called the \emph{birth} and \emph{death} values of the corresponding topological feature in the filtration. The translation between a one dimensional space and its sub level set (birth, death) pairs is particularly easy to visualize (Fig.\ref{fig:sublevel_pd}).

Persistent homology became an effective way to quantify changes in topology across scales when it was recognised that by adding points in $\Delta = \{(x,x)|x\in\mathbb{R}\}$ with countably infinite multiplicity to the multiset of (birth, death) pairs arising from the process above, the resulting persistent homology \emph{barcode} or \emph{persistence diagram} (PD) is a stable representation under perturbations to $f$ \cite{Cohen-Steiner2007}. In other words there is then a natural metric on PDs bounded above by some distance between the underlying functions \cite{dAmico2003,Ghrist2008a,Vejdemo-Johansson2012}. Frequently this stability is stated as a bound on the $p$-Wasserstein distance between sublevel set PDs of Lipschitz functions with respect to the $L_\infty$ metric: $$d_W^p(X,Y) := \left( \inf_\gamma \sum_{x\in X} \|x-\gamma(x)\|^p_\infty \right)^\frac{1}{p} \leq C\|f-g\|_\infty^d$$ where diagrams $X,Y$ are PDs for $f,g$, the constants $C,d$ are independent of $f,g$, and $\gamma$ ranges over all bijections between the countably infinite multisets $X,Y$ (see \cite{Cohen-Steiner2010} for details). Because of this the Wasserstein distance has been at the heart of applications of TDA as it has developed. However due to the complexity of optimizing over bijections \cite{Kerber2016a} and because it imposes a complicated geometry on the space of diagrams \cite{Turner2013} there has been a glut of alternative vector space representations.

One such representation is the discrete persistence image (PI) which in turn is defined in terms of a continuous persistence surface (PS) arising from a PD \cite{Adams2017}. The latter is a way to estimate the distribution from which a diagram is sampled while ensuring that the estimate itself is stable. Given a diagram $D\in \mathbb{R}_{y\geq x}$ as defined above, note that the persistence value $p:=d-b$ is at least zero for all points so we can translate $D$ to the positive quadrant via $T:(b,d)\mapsto(b,p)$. Next consider a weighted sum of Gaussians centered at points $x\in T(D)$: $$\rho_D(z) := \frac{1}{2\pi\sigma^2}\sum_{x\in T(D)} f(x)e^{-\frac{\|z-x\|^2}{2\sigma^2}},$$ where $f:\mathbb{R}^2\to \mathbb{R}$ is a continuous, piecewise differentiable function that decays to zero on the horizontal axis. This is the \emph{persistence surface} associated to $D$. It is a stable estimate of the distribution underlying $D$ given the constrains on $f$ (see \cite{Adams2017,Chazal2018} for details). To provide a vector space representation of $\rho_D$ suitable for use in machine learning applications, a finite regular grid $G$ is then placed over part of $\mathbb{R}^2_+$ and used to quantize $\rho_D$ as a collection of pixels $I(\rho_D)_g:=\iint_g \rho_D(z) \,dz$, corresponding to each cell $g\in G$. This collection of pixels is the \emph{persistence image} representation of $D$ and it is a suitable vector representation for off-the-shelf classifiers such as support vector machines (SVMs) to use \cite{Adams2017}.

\subsection{Entropy Regularized Discrete Optimal Transport}

The question of optimizing transport costs between distributions of resources has been studied in various forms since the 1700s, and in its modern guise is central to statistical learning theory \cite{Villani2003,Villani2008}. Given two $n$-bin histograms $r,c$, a \emph{transport plan} between $r$ and $c$ is a matrix $P\in\mathbb{R}_+^{n\times n}$ satisfying $\sum_i p_{ij} = r$ and $\sum_j p_{ij} = c$. Equivalently, $P$ is a joint probability $P=P(X,Y)$ for two multinomial random variables $X$ and $Y$ taking values in $\{0,1,\ldots,n-1\}$, whose marginals are $r$ and $c$. We write $U_n(r,c)$ for the set of all transport plans for $n$-bin histograms $r,c$, dropping the subscript $n$ when it is clear from context.

Given a transport plan $P$, we interpret $p_{ij}$ as a mass to be transported from the $i$-th component of $r$ to the $j$-th component of $c$. If the cost of this operation is $m_{ij}\in \mathbb{R}$ per unit of mass transported, then the \emph{discrete optimal transport \emph{(OT)} problem} is to minimise the sum of transport costs over all possible plans for $r,c$ given $M$: $$OT_M(r,c) := \min_{P\in U(r,c)}\langle P, M \rangle$$ where $\langle P,M \rangle$ is the Frobenius product $\sum_{i,j=1}^n p_{ij}m_{ij}$. The function $OT_M$ is a metric on the space of histograms when $M$ is itself a metric distance matrix \cite{Avis1980,Villani2003} but computing it exactly is difficult in practice, with the worst case time complexity $O(n^3 \log n)$ being reached with certain values of $r,c$ and $M$ \cite{Pele2009}.

Recent work showed that regularizing the classical $OT_M$ problem by adding a convex constraint can lead to fast approximations \cite{Altschuler2017,Benamou2015,Blondel2017,Cuturi2013a,Dessein2016,Solomon2015}. Define the entropy of a transport plan $P$ as $H(P) := -\sum_{i=1}^n p_{ij}\log p_{ij}$, then minimising the sum of transport costs over \emph{high entropy} transport plans as in $$ROT_M^{\lambda\geq0}(r,c) := \langle P^\lambda, M\rangle$$ where $$P^\lambda :=\mathrm{argmin}_{P\in U(r,c)} \left( \langle P,M \rangle - \lambda H(P) \right)$$ gives an upper bound approximation to $OT_M$ that has complexity $O(n^2 \log n)$ \cite{Altschuler2017}.\footnote{Different penalty functions lead to approximations with different convergence characteristics \cite{Blondel2017}, but we consider only entropy in this paper.} This upper bound is called the \emph{Sinkhorn divergence} between the histograms since it has a natural parallel implementation based on iterated matrix-vector products known as the Sinkhorn Knopp (SK) algorithm \cite{Cuturi2013a}. Early evidence suggests that it gives better classification results than $OT_M$ in a number of experiments and that it converges very quickly in practice \cite{Altschuler2017}. Moreover, when the cost matrix $M$ is highly structured as is the case for $L_p$ distances on regular grids, the matrix operations of the algorithm can be speeded up further via FFT based convolutions \cite{Solomon2015}. This speed and accuracy advantage has increased interest in Sinkhorn divergence and other regularized variants of $OT_M$ for a variety of problems ranging from color transfer in image processing to model optimization in machine learning \cite{Blondel2017,Genevay2017}. In the following sections we show that it can be integrated successfully within classification pipelines based on topological features as well.

\section{Sinkhorn Divergence of Topological Signature Estimates}
\label{sec:method}

The problem we address now is to classify time series generated by deterministic dynamical systems. Suppose we are given time series samples from two classes corresponding to a choice of parameters in a single dynamical model. Can we characterize the data in terms of their shared or distinct topological properties without resorting to embeddings or parametric methods, and use this to effectively predict the class labels of new series? Represent a dataset containing $m$ samples each of length $n$ using an array $X=(x_i^j)_m^n$, and represent class labels using a 0-1 vector $y=(y_i)_m$.\footnote{The model we define is equally valid for classifying time series of different lengths and with more than two classes, with minor adjustments.}

\subsubsection*{Training}
\begin{enumerate}[label=T\arabic*]
	\item \label{training:persistence} For each sequence $x_i$ directly compute its sublevel set persistence diagram $D_i$.

	\item \label{training:partition} Partition the set of $m$ PDs according to their associated class labels in $y$, giving two sets of PDs: $D^\dag = \{D_i\,|\,y_i = 0\}$ and its compliment $D^\ddag = \{D_i \,|\, y_i=1 \}$.

	\item \label{training:overlay} For each of $D^\dag$ and $D^\ddag$ overlay the points in its member diagrams to form a combined persistence diagram representing the whole class: $\overline D^\dag = \bigcup D^\dag$ and $\overline D^\ddag = \bigcup D^\ddag$.

	\item \label{training:param_choice} Choose a continuous and piecewise differentiable function $f:\mathbb{R}^2_+\to \mathbb{R}$ such that $f(x,0)=0$ for all $x$, and a smoothing radius $\sigma$. Construct the smoothed persistence surfaces $\rho_{\overline D^\dag}$ and $\rho_{\overline D^\ddag}$.

	\item \label{training:grid} Choose a $d\times d$ square grid $G$ that extends beyond the largest values of $b$ and $p$ in $\overline D^\dag \cup \overline D^\ddag$. Compute the persistence images $I(\rho_{\overline D^\dag})$ and $I(\rho_{\overline D^\ddag})$ over the cells of $G$.
\end{enumerate}
In practice the form of $f$, the value $\sigma$, and the size of $G$ can all be set at this stage using cross validation on the training data. After stage \ref{training:grid} we have for each class a stable kernel estimate of the density of its expected persistence diagram, which naturally leads to the following prediction pipeline.

\subsubsection*{Prediction}
\begin{enumerate}[label=P\arabic*]
	\item \label{prediction:persistence} Given an unlabelled query sequence $q$ compute its persistence image $I_q$ using the pipeline above but skipping the diagram overlay steps \ref{training:partition} and \ref{training:overlay}, and reusing the same values for $f$, $\sigma$ and $G$ chosen in \ref{training:param_choice}.

	\item \label{prediction:sinkhorn} Choose a $p$ value for an underlying $L_p$ metric on the grid $G$, which induces a cost matrix $M_p$ on $G$. Also choose a regularization parameter $\lambda\geq 0$ for computing Sinkhorn divergences $ROT_{M_p}^\lambda$. Compute the Sinkhorn divergences $$d^\dag = ROT_{M_p}^\lambda(I_q, I(\rho_{\overline D^\dag})),\qquad d^\ddag = ROT_{M_p}^\lambda(I_q, I(\rho_{\overline D^\ddag})).$$
	
	\item \label{prediction:predict} If $d^\dag < d^\ddag$ then predict $y=0$, if $d^\ddag < d^\dag$ then predict $y=1$, else predict $y=0$ or $y=1$ with equal probability.
\end{enumerate}
In practice the values of $p$ and $\lambda$ can both be optimized using cross validation during the training phase.

Thus our model predicts labels for new time series based on the closest expected persistence diagram for each class in the training set, using the entropy regularized optimal transport distance between the distributions.

\subsubsection*{Implementation}
Computing the sub level set persistence of each time series at stages \ref{training:persistence} and \ref{prediction:persistence} requires determining its critical points (local maxima and minima) and also noting for each local maximum which of its two neighbouring minima is closer in value. Thus the critical points must be sorted as part of the process, which is an $O(n\log n)$ operation at worst and $O(mn\log n)$ for each class.

To compute the values of the persistence image pixels $I_g(\rho_{\overline D^\dag})$ and $I_g(\rho_{\overline D^\ddag})$ for $g\in G$ various numerical integration and approximation methods are available. In particular if we assume that each point appearing in a cell is centered in that cell we can approximate the persistence surfaces $\rho_{\overline D^\dag}$ and $\rho_{\overline D^\ddag}$ by convolving their underlying $d\times d$ $f$-weighted histograms with a discrete filter corresponding to our chosen Gaussian. This allows us to compute the persistence images generated by \ref{training:param_choice} and \ref{training:grid} in a single step, in $O(d^2 \log d)$ for our grid.

Finally during prediction, computing the regularized optimal transport cost between two $d^2$-bin histograms for a cost matrix $M_p$ corresponding to $L_p$ distances on the grid is $O(d^2 \log d)$. This is because $M_p$ is a block Toeplitz of Toeplitz blocks (BTTB) matrix in this situation, meaning the matrix-vector products appearing in the Sinkhorn-Knopp (SK) algorithm used to compute $ROT_{M_p}$ can be computed using FFT enhanced convolutions. See \cite{Altschuler2017,Cuturi2013a} for details of the SK algorithm and in paricular Chapter 5 of \cite{Vogel2002} for details of how to speed up the matrix-vector operations.

The result is that once the size $d$ of the grid has been set during training the complexity of the model is $O(n\log n)$ in time series length.

\section[Classification Experiments]{Classification Experiments\footnote{Sklearn-compatible Python code implementing the classifiers described here can be found at https://github.com/colinstephen/icmla2018}}
\label{sec:experiments}

We call the method above `Persistence Image Classification using Regularized Optimal Transport', or \classifier for short. This section assesses its performance against one classifier using signal frequency and rate of change analysis, and one classifier based on persitent entropy as defined and successfully applied to similar problems in \cite{Rucco2015}.\footnote{General purpose time series classifiers such as those benchmarked in \cite{Bagnall2017} do not seem to perform well for the dynamic systems considered here. Initial results using \textit{dynamic time warping} (DTW) and \textit{random forests} were not competitive in terms of accuracy, while the potentially high-performance \textit{collective of transformation ensembles} (COTE) and the \textit{elastic ensemble} (EE) methods were too slow to evaluate due to the lengths of time series used here.}

\begin{enumerate}[label=C\textsub{\arabic*}]
	\item \classifier compares kernel estimates of PD densities using the Sinkhorn divergence. We fix a weight function for \classifier that increases rapidly from zero to one in an interval less than the persistence value of any off diagonal points processed. Thus in effect we apply uniform weights when constructing persistence images. The smoothing parameter $\sigma$ and grid size $d$ for PIs, and the regularization parameter $\lambda$ for the Sinkhorn metric, are all estimated using 5-fold cross validation over a grid of candidate values during training.
	
	\item CEPS is the one nearest neighbor classifier using Euclidean distance between coefficients of the discrete cosine transforms of the cepstra of the time series being compared: $$d_\textrm{CEPS}(T_1, T_2) := \left( \sum_i | \textrm{CEPS}(T_1)_i - \textrm{CEPS}(T_2)_i |^2 \right)^\frac{1}{2}$$ where $$\textrm{CEPS}(T) := \textrm{DCT}\left( \left| \mathcal{F}^{-1} \left\{ \log ( |\mathcal{F}(t)|^2) \right\} \right|^2 \right),$$ $\mathcal{F}$ is the Fourier transform, and the sum is over all coefficients

	\item PENT is the one nearest neighbor classifier using the absolute difference between persistent entropies \cite{Rucco2017}. If $D(T)=\{ (b_i, d_i) \,|\, i\in I \}$ is an indexed set of the off-diagonal points in the persistence diagram associated to $T$, $p_i:=(d_i-p_i)$ is the persistence of each point and $P:= \sum_i p_i$ is the total persistence of the diagram, then $$d_\textrm{PENT}(T_1,T_2) := |\textrm{PENT}(T_1) - \textrm{PENT}(T_2)|$$ where $$\textrm{PENT}(T) := -\sum_{i} \frac{p_i}{P} \log \left( \frac{p_i}{P} \right).$$ 
\end{enumerate}

CEPS is not commonly used as a general time series similarity measure but it is effective when the series have an underlying regularity or cyclicity \cite{Kalpakis2001,Randall2017} as with the problems here. The measure $d_\textrm{CEPS}$ captures information about the relative rates of change of the two signals across their frequency bands. Cross validating the number of leading terms compared in the cepstral classifier did not improve results, so we compare the entire cepstra. On the other hand $d_\textrm{PENT}$ compares signals using a coarse grained statistic derived from their persistence diagrams. We also use one nearest neighbour here because it shows higher accuracy than the receiver operating characteristic (ROC) optimized threshold approach originally appearing in \cite{Rucco2017}.

\subsubsection*{Data}

Binary time series classification problems were generated using the combinations of parameter values for the systems defined in Table \ref{table:systems} (13 parameter value combinations). The parameter ranges were chosen to ensure that all trajectories studied here are chaotic. Initial conditions and parameter values themselves were varied uniformly in the given intervals to ensure a wide variety of trajectories were observed.

In the case of the Henon and Lorenz systems only the $x$ values were used. In all cases the raw time series were $z$ normalized before any training or predictions and the first 1,000 sequence values were discarded. The subsequent lengths used for classification varied from 2,500 to 15,000 in steps of 2,500 (6 lengths total). White noise was added to each sample varying in standard deviation from 0.0 (no noise) to 0.75 in steps of 0.125 (7 noise levels in total). Thus there were 546 classification experiment configurations in total. Two example time series snippets from different classes in the fourth Henon configuration are shown in Figure \ref{fig:series_snippets}.

For each of the 546 configurations 200 samples balanced between the two classes were generated, giving a total of 109,200 time series to be processed on each run. The data for each experiment were split randomly in to 170 training and 30 test time series to be classified. This process was then repeated for 10 runs per experiment, using different random train-test splits each time.
 
\begin{table*}[htb]
\begin{center}
\caption{Parameters used to generate time series.}
\label{table:systems}
\begin{tabular}{|c|c|c|c|}\hline
System & Model & Class 0 & Class 1 \\ \hline

\multirow{4}{*}{Logistic} & \multirow{4}{*}{
	\makecell{
		$x_{n+1} = ax_n(1-x_n)$ \\
		$x_0 \in (0,1)$ uniform
	}
} & \multirow{4}{*}{$a = 3.9995 \pm0.0005$} & $a = 3.9945\pm0.0005$ \\
& & & $a = 3.9895\pm0.0005$ \\
& & & $a = 3.9845\pm0.0005$ \\
& & & $a = 3.9795\pm0.0005$ \\ \hline

\multirow{5}{*}{Henon} & \multirow{5}{*}{
	\makecell{
		$x_{n+1} = 1 - ax_n^2 + y_n$ \\
		$y_{n+1} = bx_n$ \\
		$x_0, y_0 \in (1,2)$ uniform
	}
} & \multirow{5}{*}{
	\parbox{0.1cm}{
	\begin{align*}
	a &= 1.4\pm0.0025 \\
	b &= 0.3035\pm0.0025
	\end{align*}
	}} & $b=0.3085\pm0.0025$ \\
& & & $b=0.3135\pm0.0025$ \\
& & & $a=1.395\pm0.0025$ \\
& & & $a=1.390\pm0.0025$ \\
& & & $a=1.385\pm0.0025$ \\ \hline

\multirow{4}{*}{Lorenz} & \multirow{4}{*}{
	\makecell{
		$\dot{x} = \sigma(y-x)$ \\
		$\dot{y} = x(\rho-z)-y$ \\
		$\dot{z} = xy-\beta z$ \\
		$x_0, y_0, z_0 \in (0,1)$ uniform
	}
} & \multirow{4}{*}{
	\makecell{
		$\beta = 8/3$ \\
		$\sigma = 10$ \\
		$\rho = 28$
	}
} & $\rho=27.75$ \\
& & & $\rho=27.50$ \\
& & & $\rho=27.25$ \\
& & & $\rho=27.00$ \\ \hline

\end{tabular}
\end{center}
\end{table*}

\subsubsection*{Results}

\begin{figure*}[htb]
\centering
\caption{Representative profiles of classifier accuracy vs noise for three system configurations. Lorenz, Henon, and Logistic systems with a fixed chosen set of parameters appear in columns 1 to 3 respectively. Time series length increases down the page. The classifier \classifier based on Sinkhorn divergences between persistence density estimates consistently outperforms the benchmarks.}
\label{fig:accuracy_profiles}
\begin{tabular}{ccc}
\includegraphics[scale=0.35]{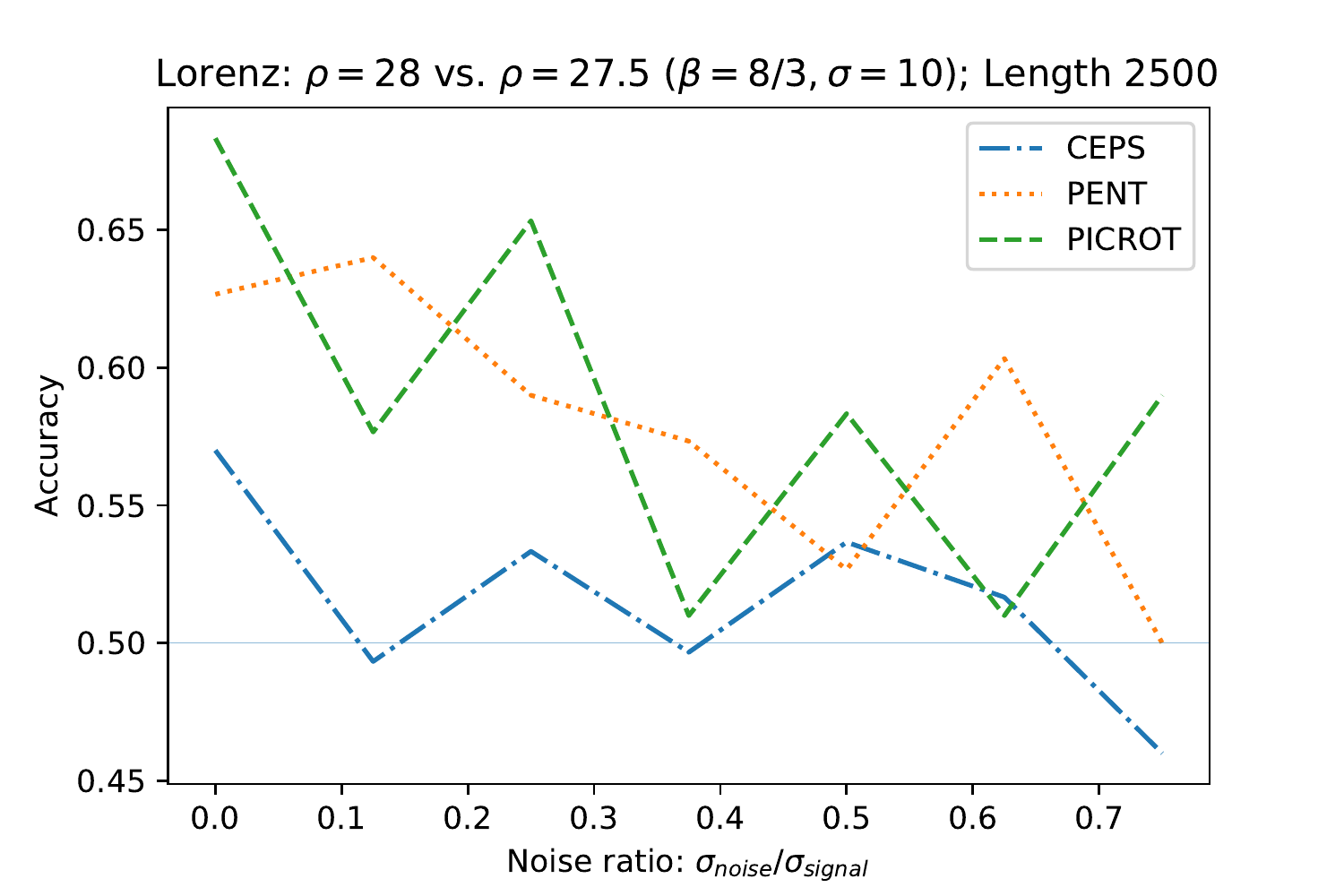} & \includegraphics[scale=0.35]{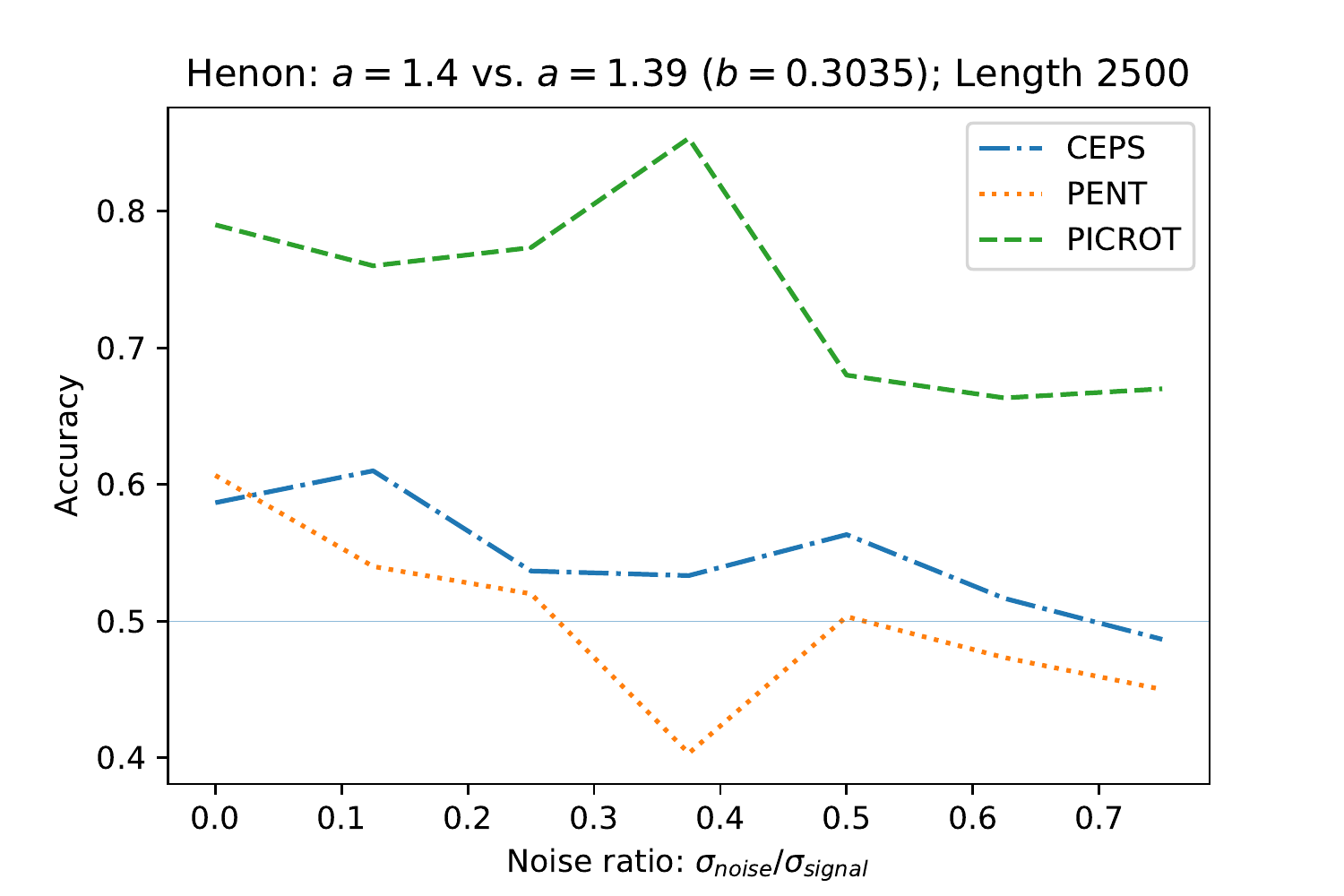} & \includegraphics[scale=0.35]{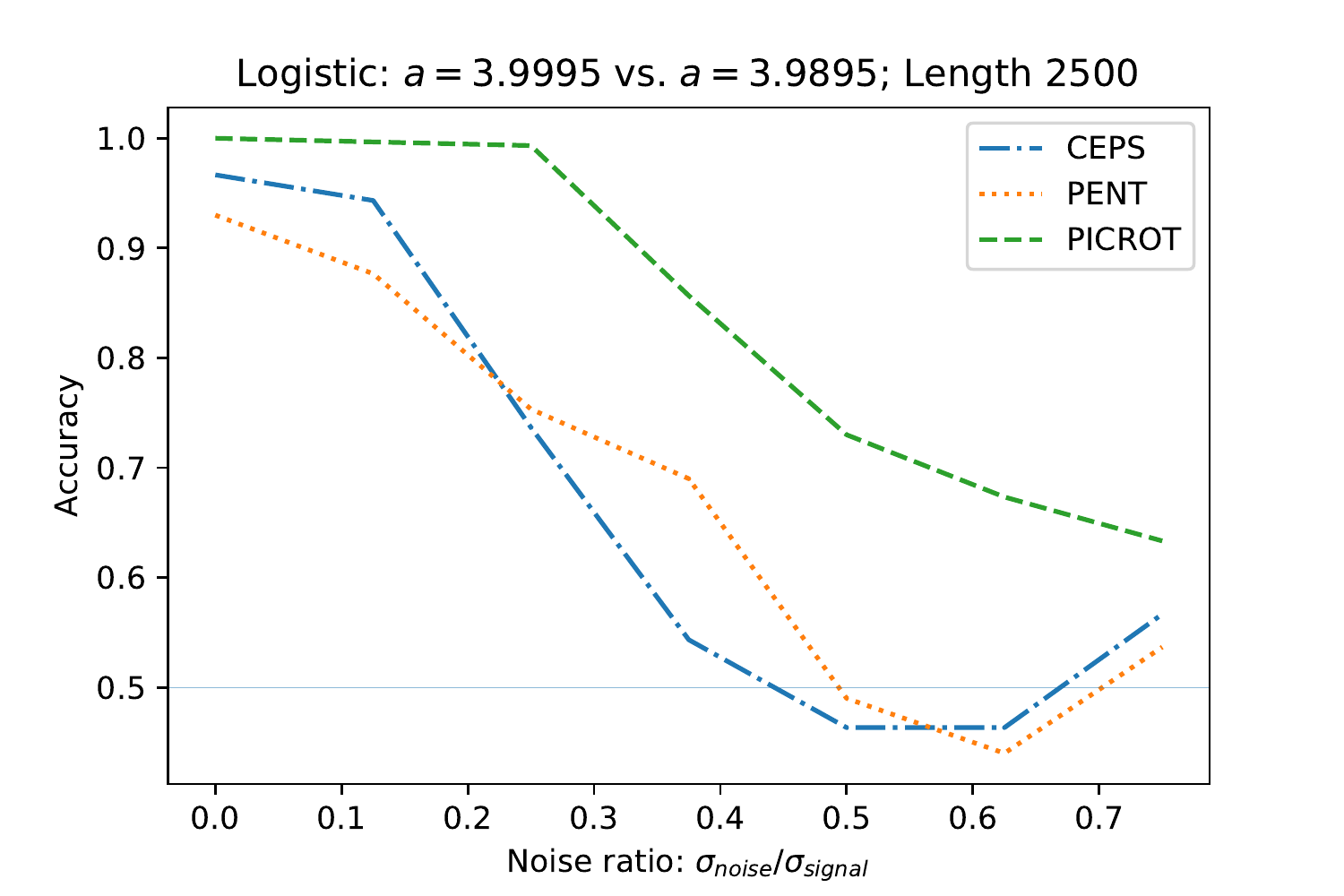} \\
\includegraphics[scale=0.35]{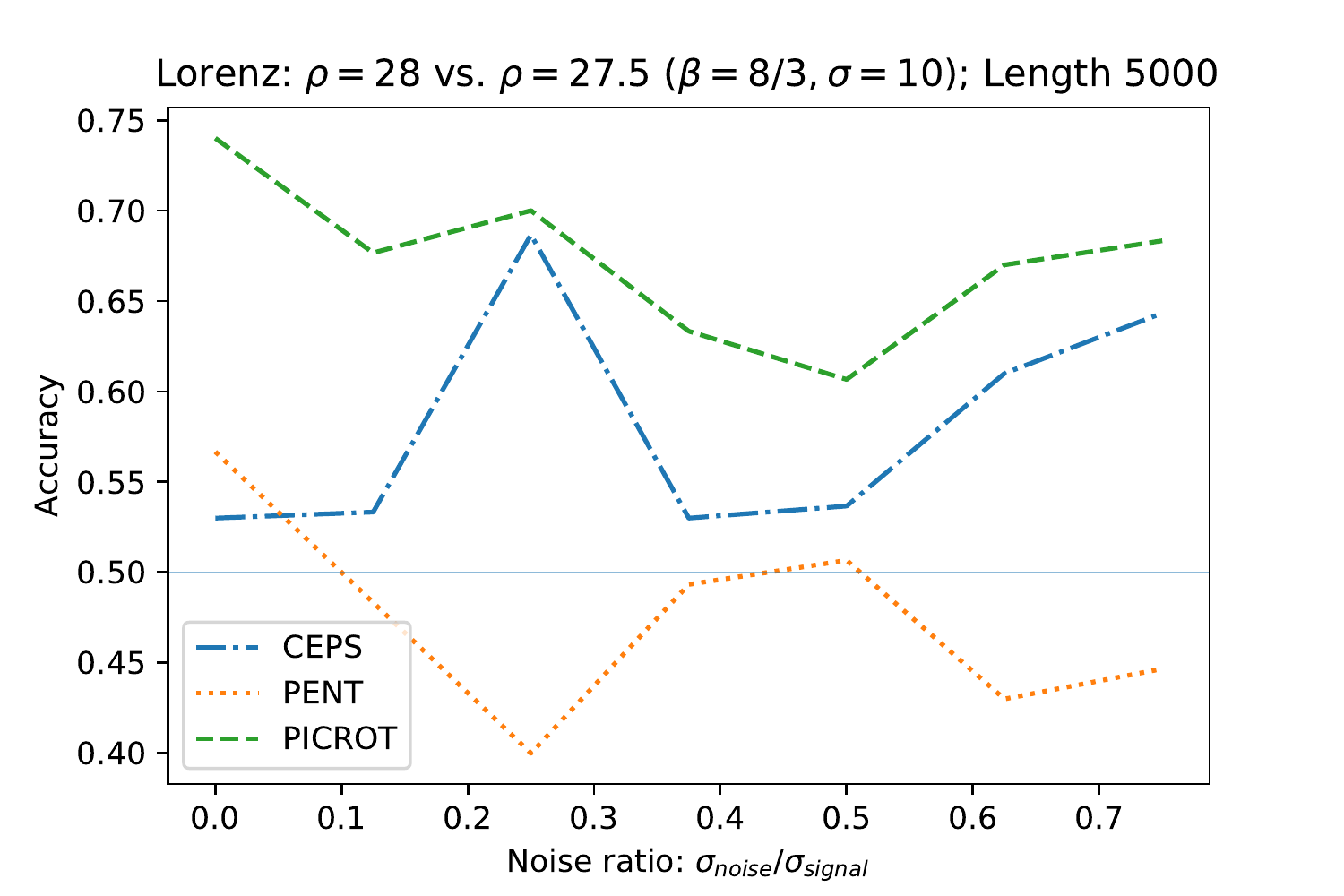} & \includegraphics[scale=0.35]{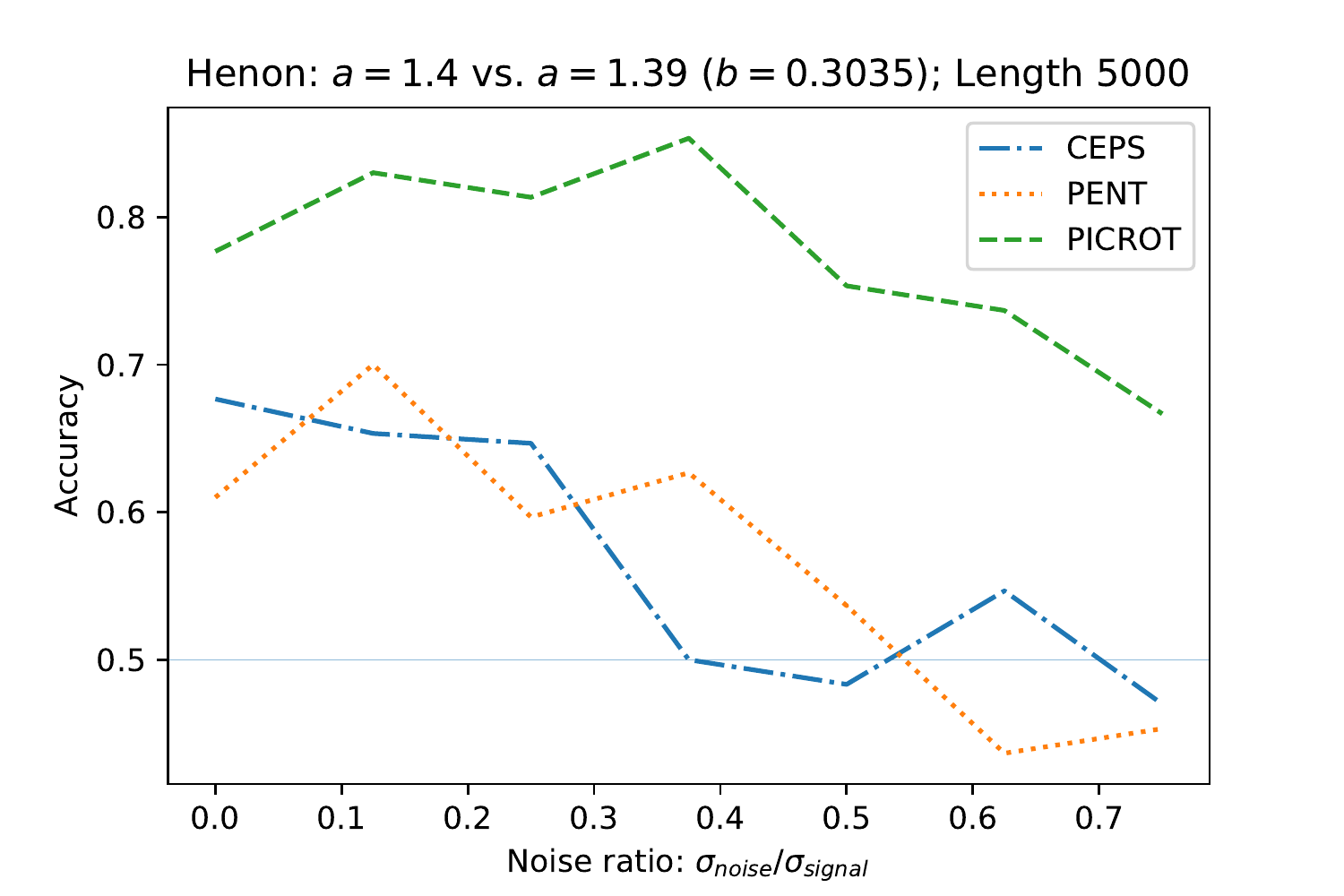} & \includegraphics[scale=0.35]{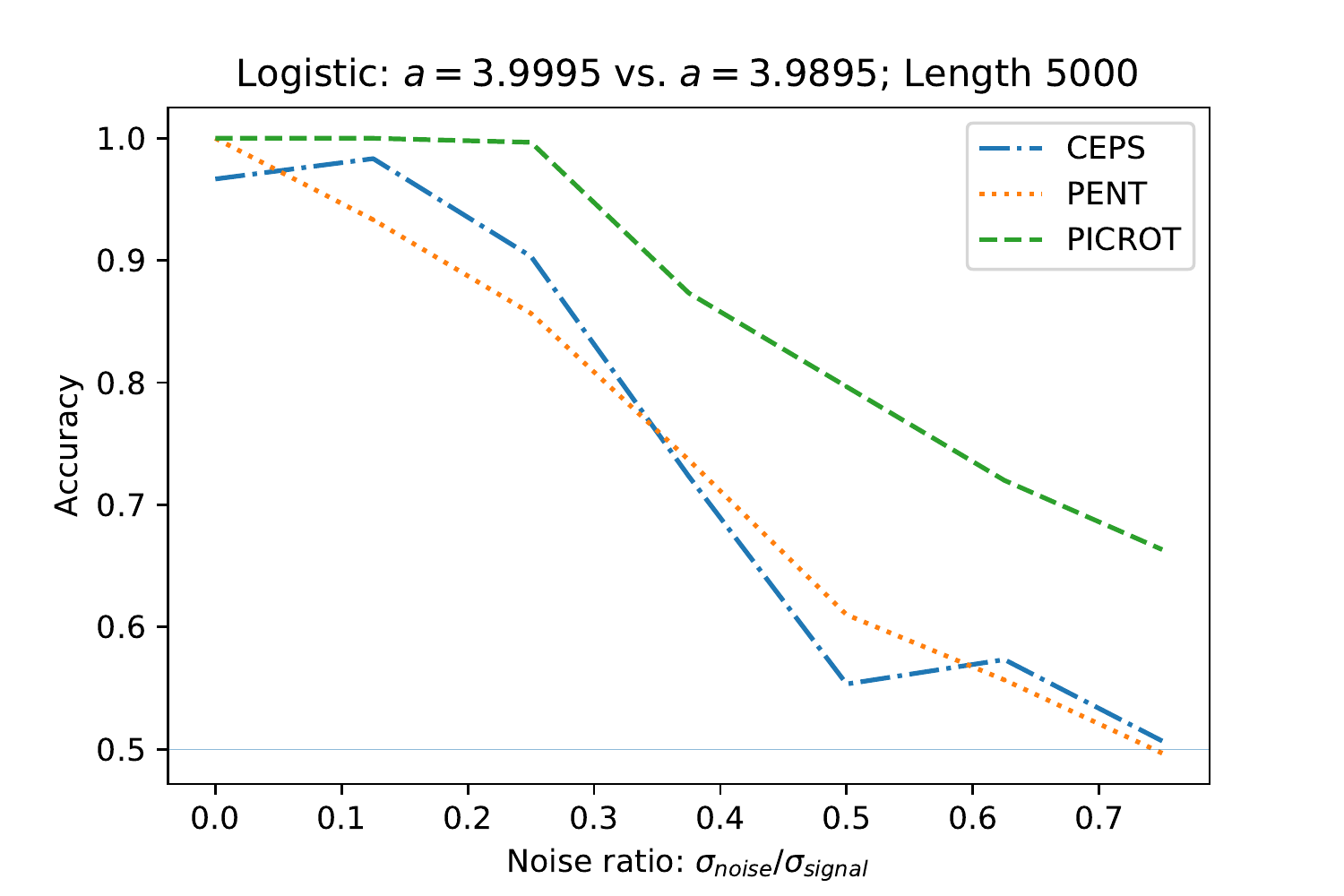} \\
\includegraphics[scale=0.35]{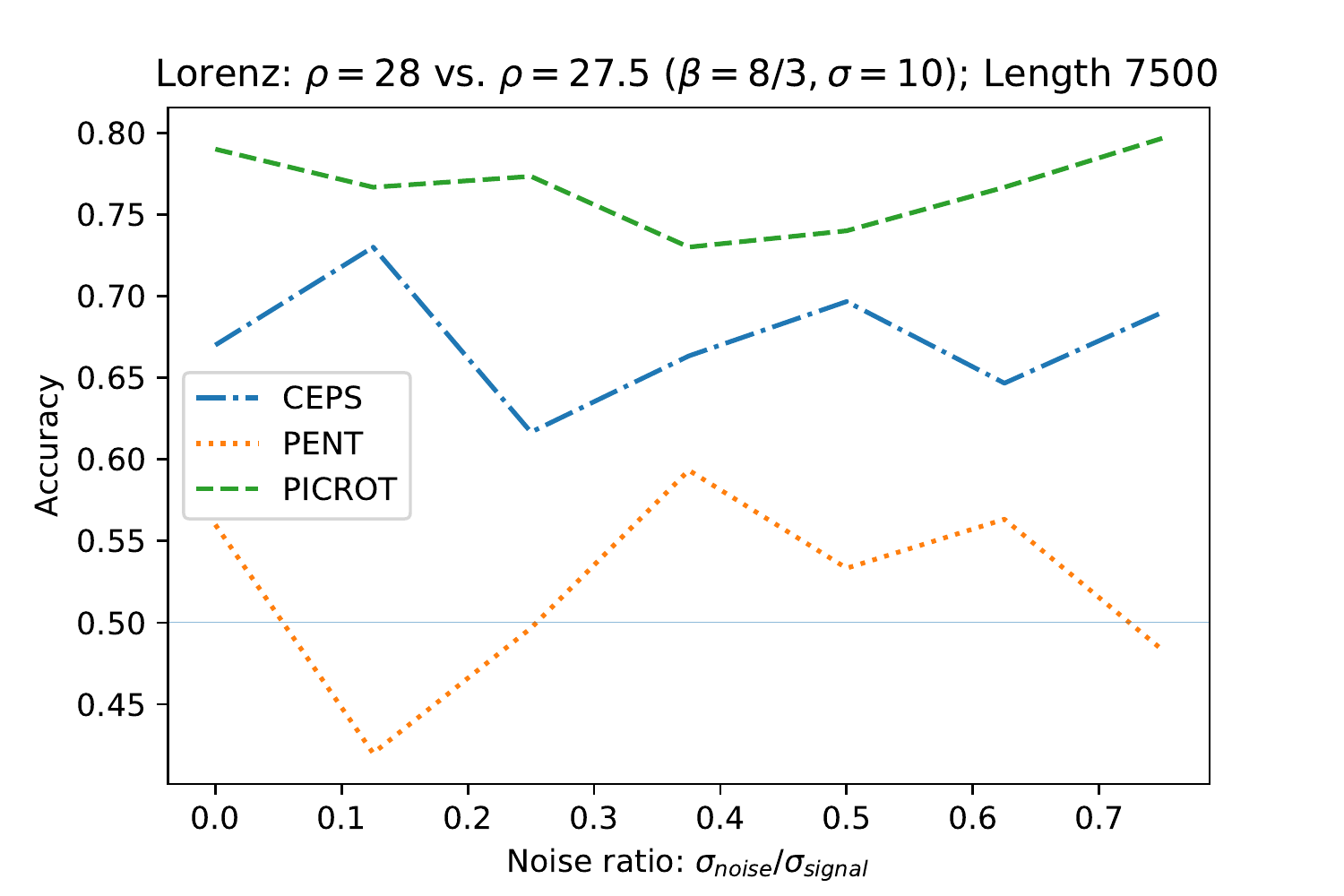} & \includegraphics[scale=0.35]{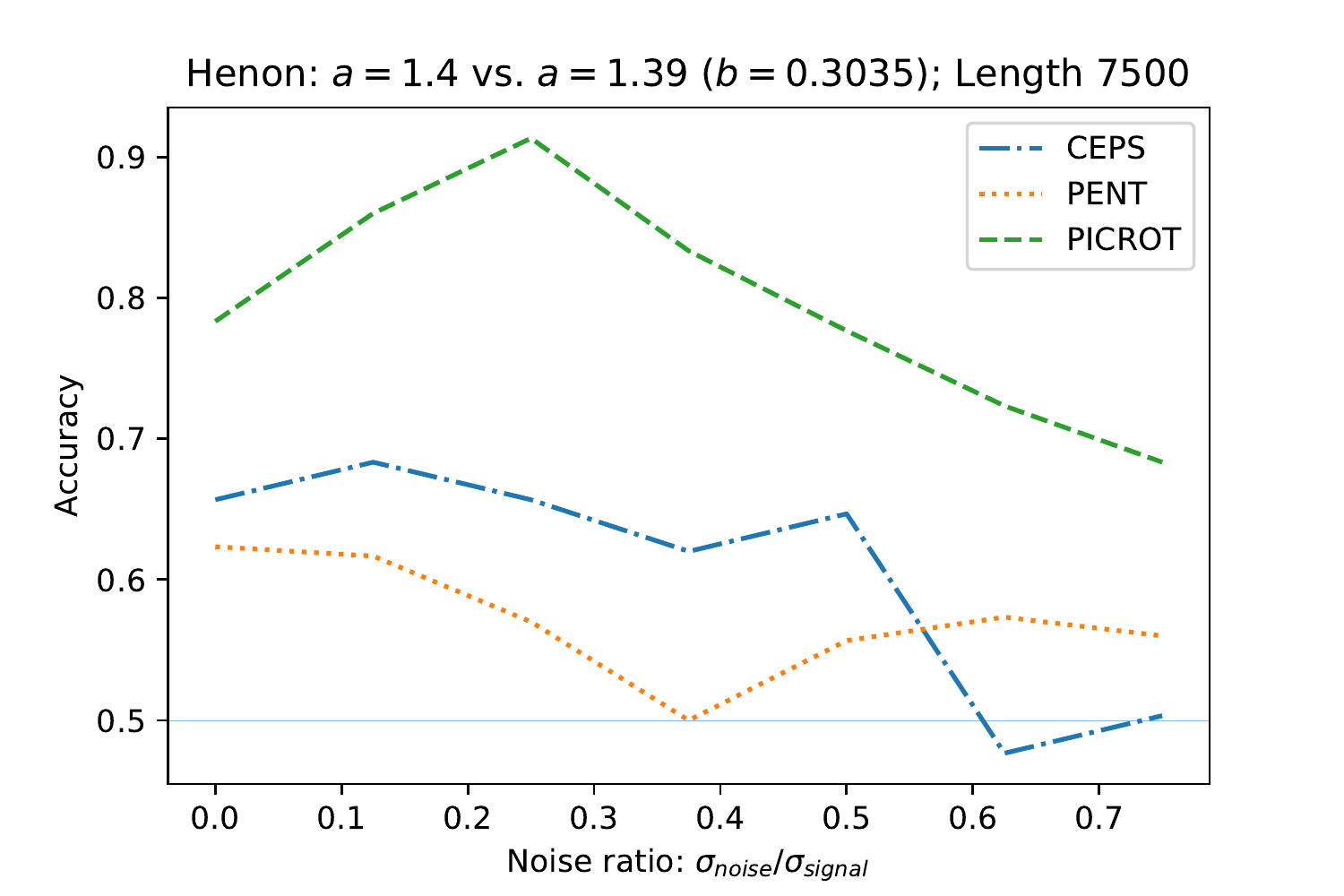} & \includegraphics[scale=0.35]{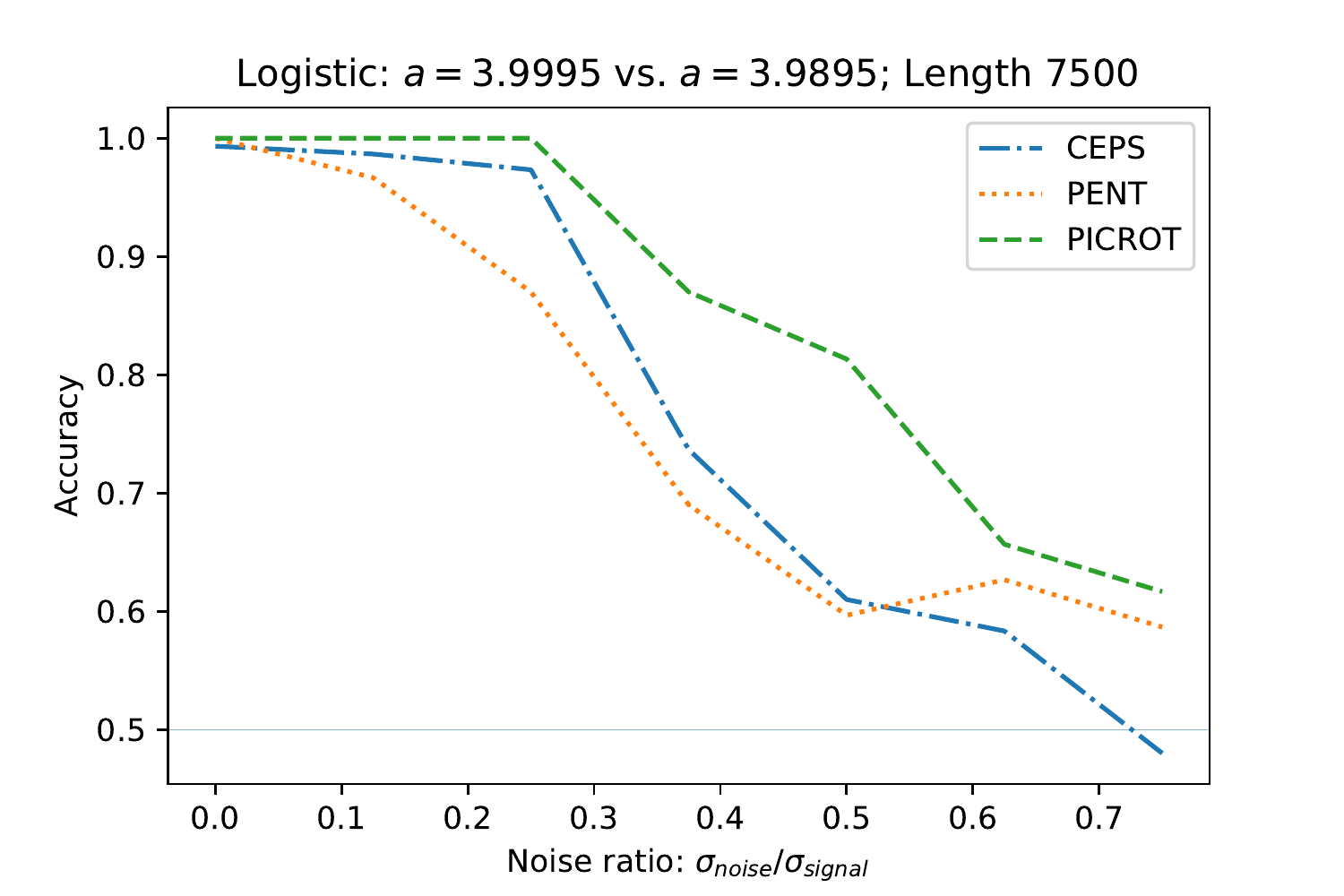} \\
\includegraphics[scale=0.35]{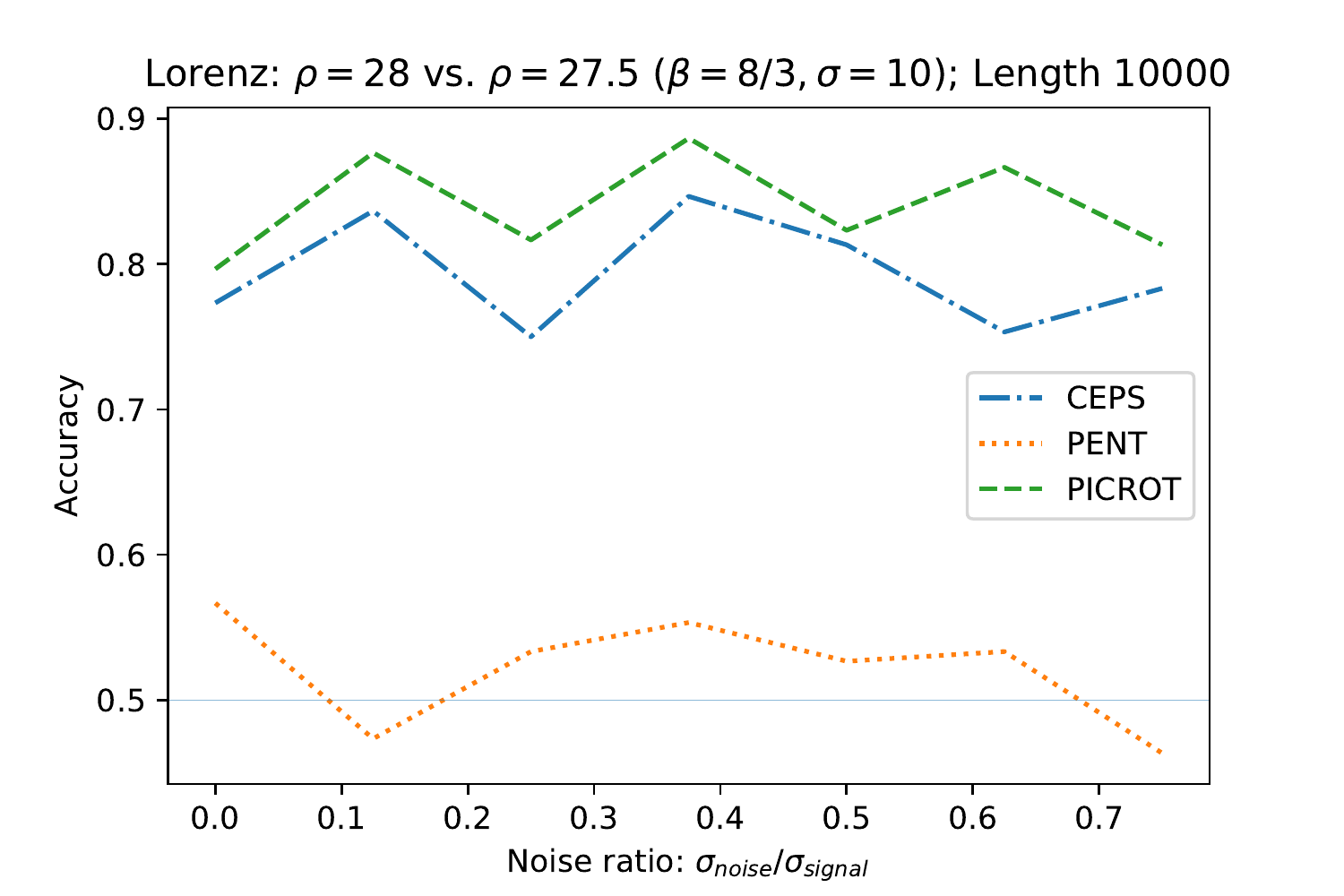} & \includegraphics[scale=0.35]{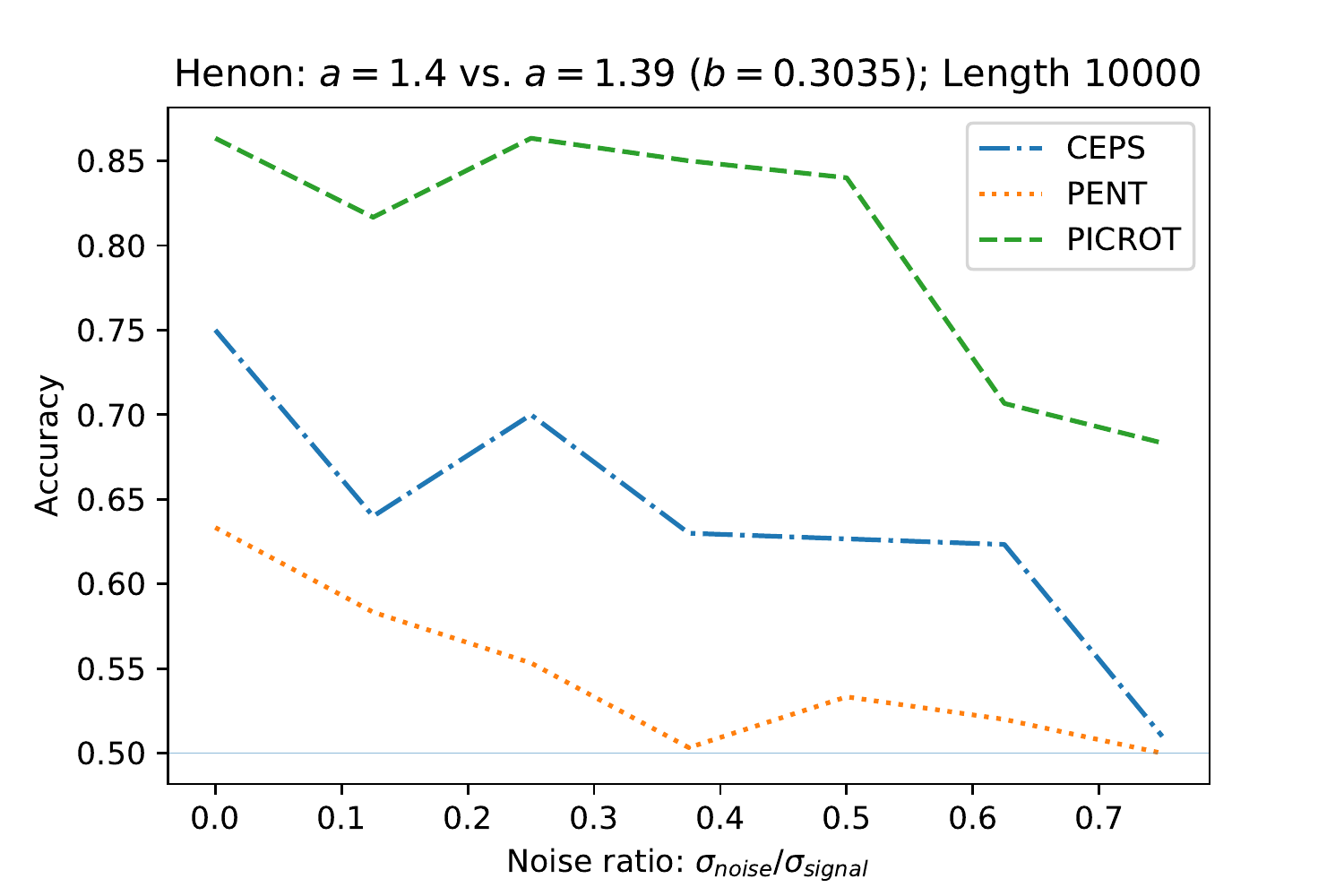} & \includegraphics[scale=0.35]{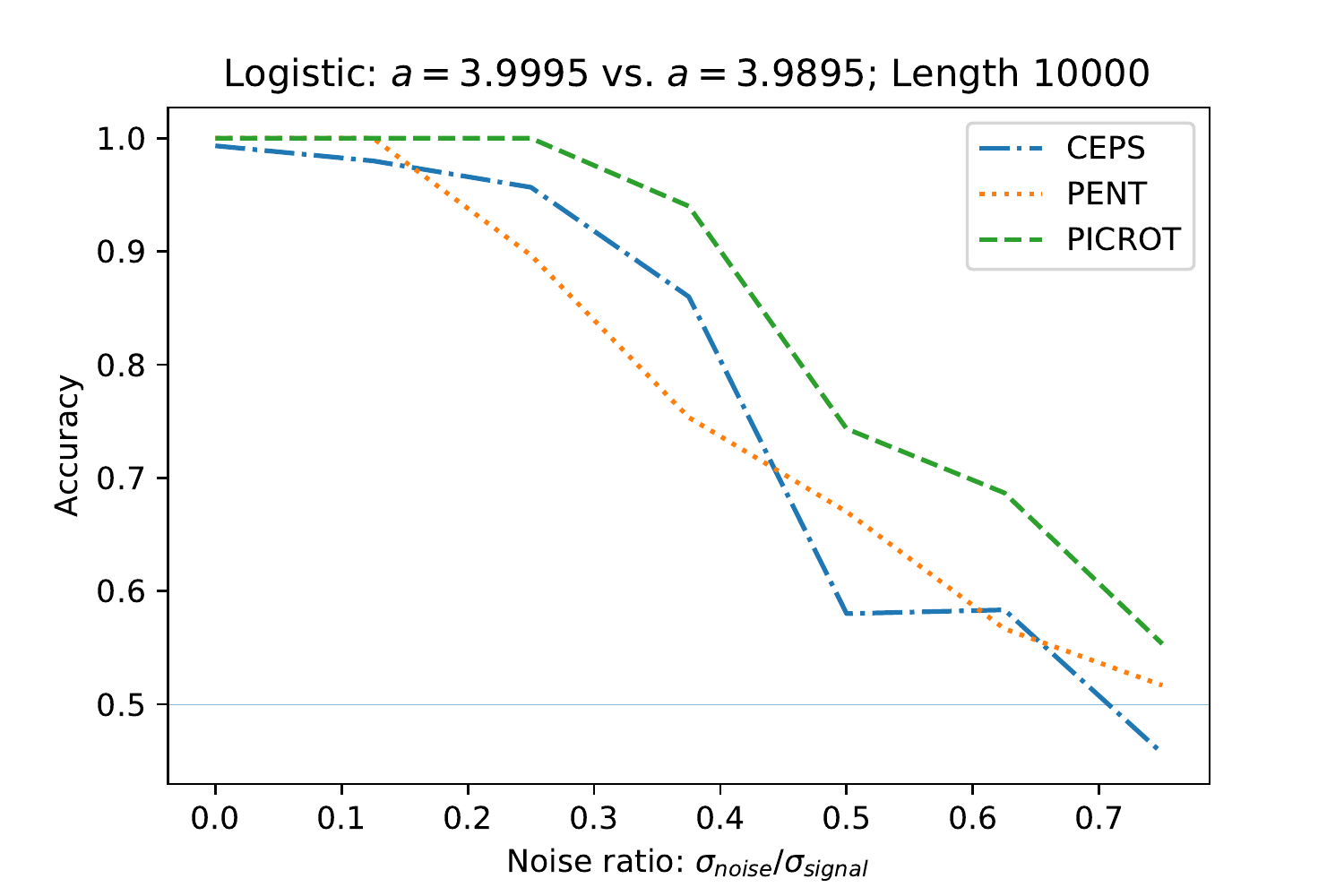} \\
\includegraphics[scale=0.35]{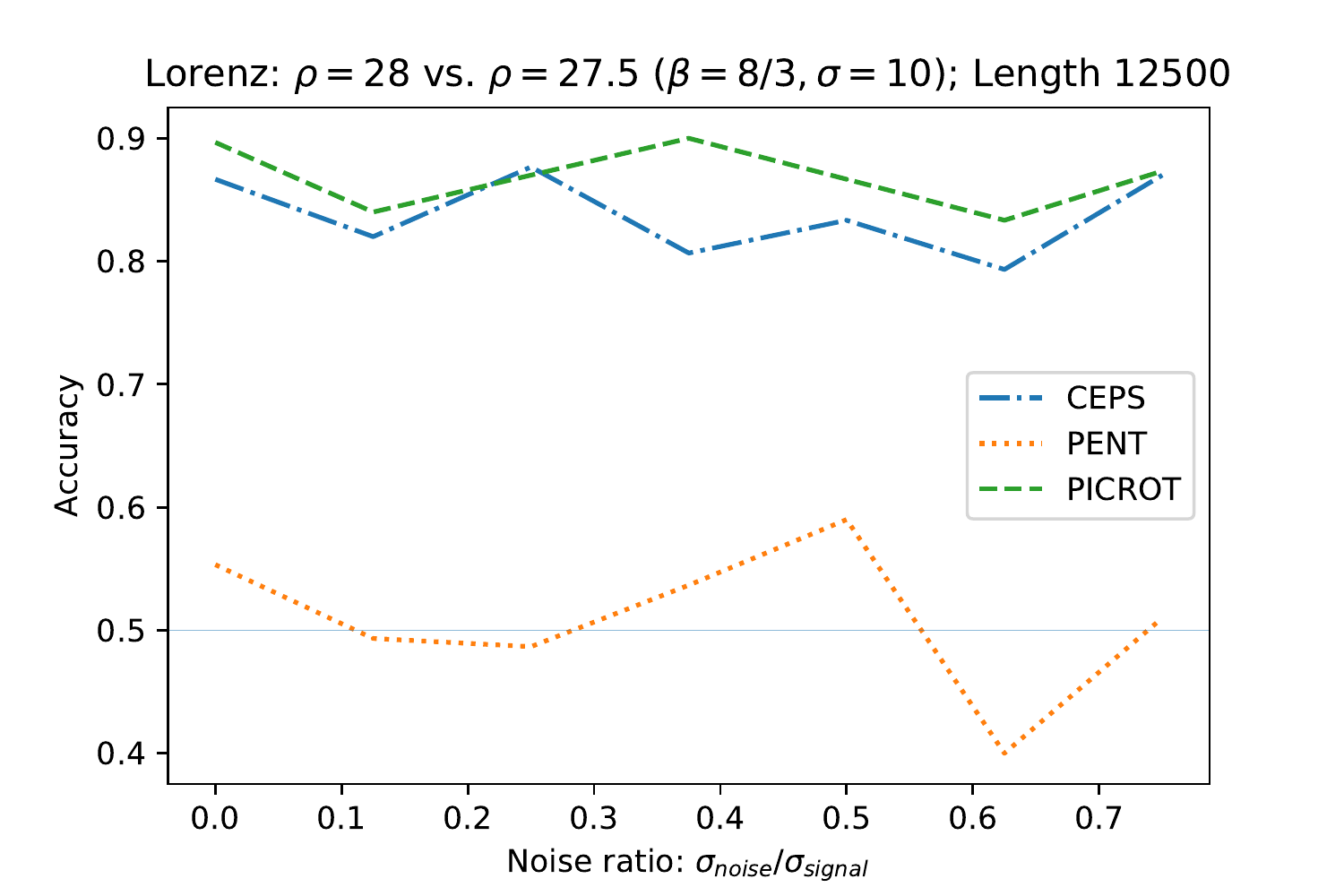} & \includegraphics[scale=0.35]{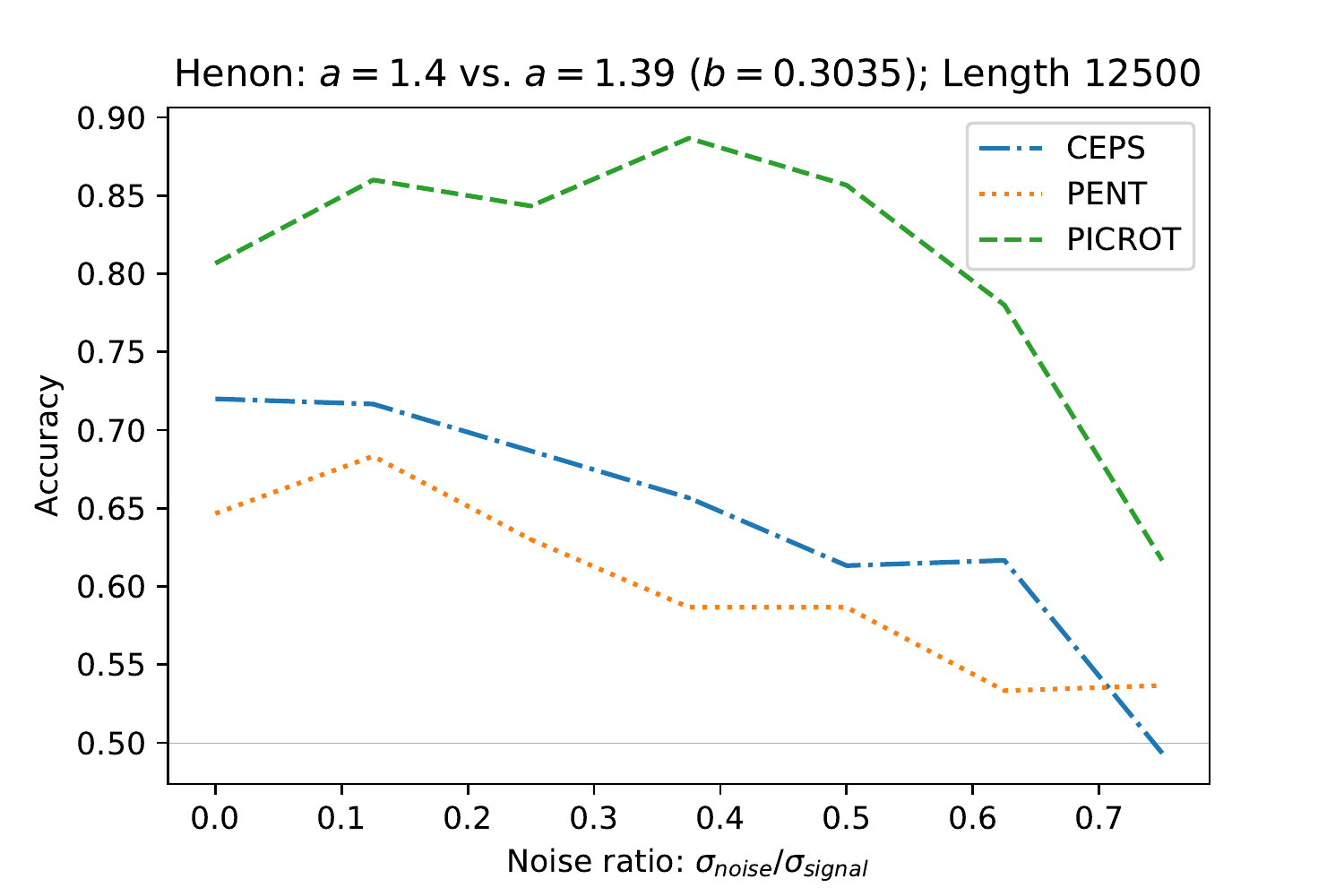} & \includegraphics[scale=0.35]{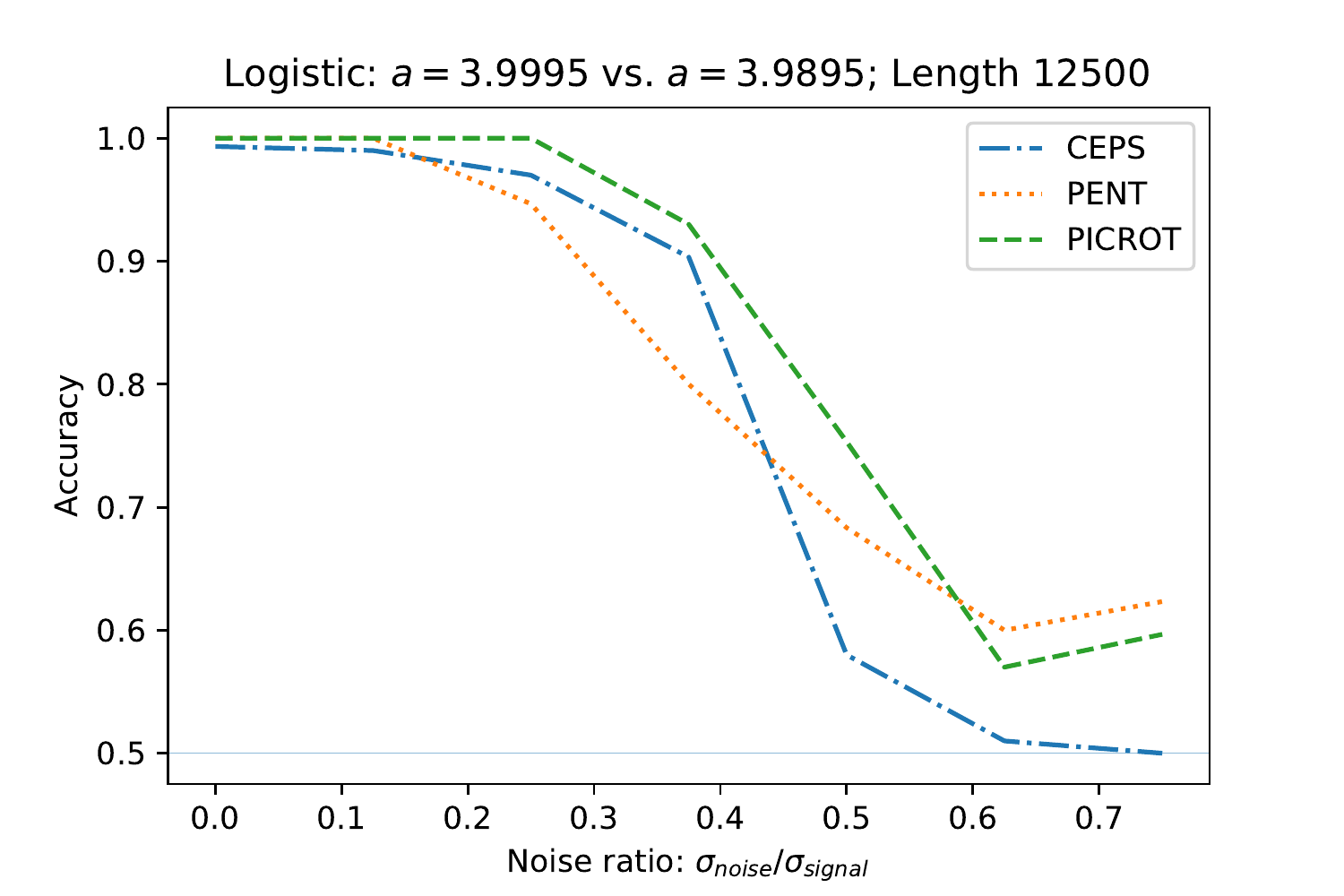} \\
\includegraphics[scale=0.35]{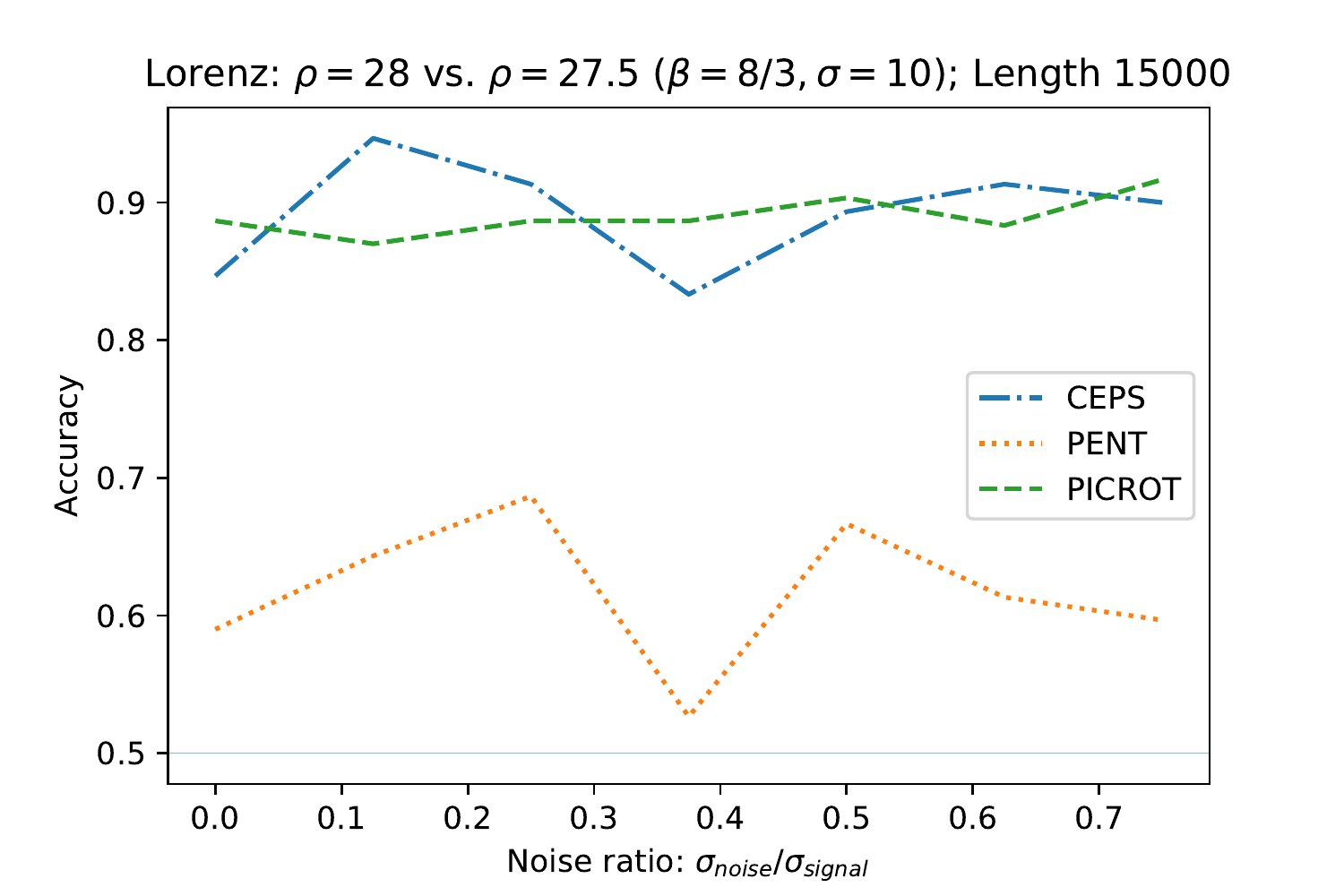} & \includegraphics[scale=0.35]{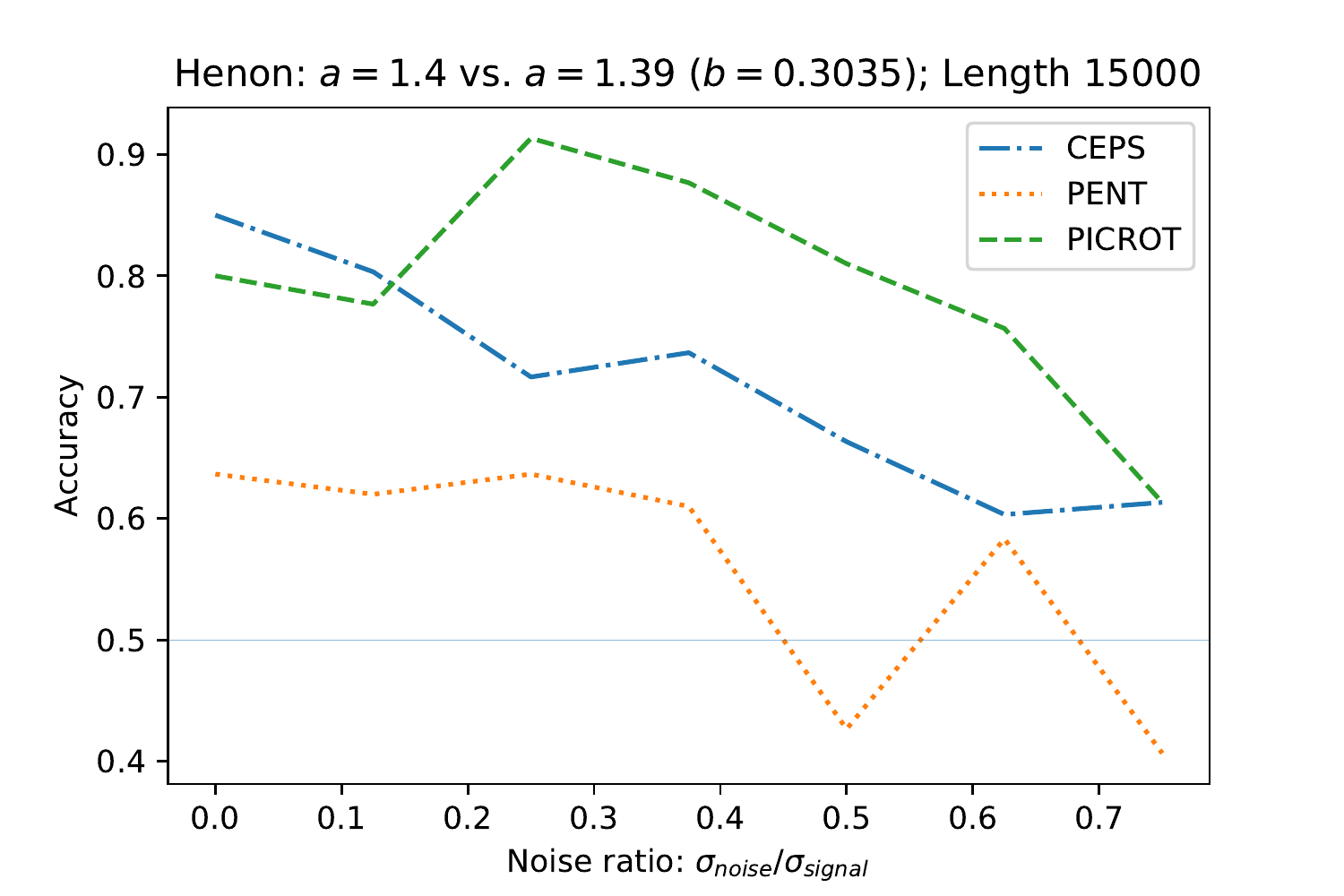} & \includegraphics[scale=0.35]{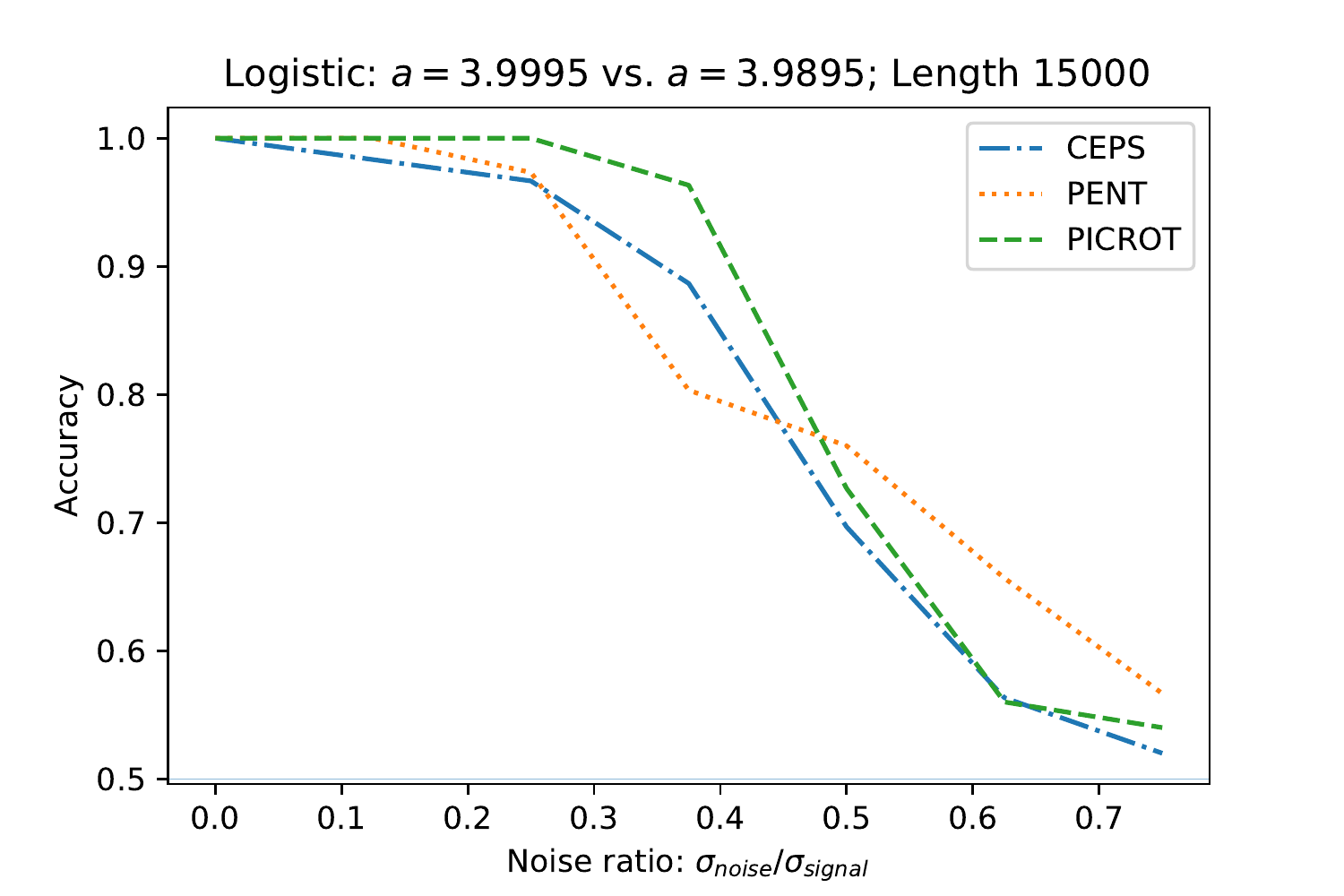} \\
\end{tabular}
\end{figure*}

\begin{figure*}[htb]
\centering
\caption{Grayscale maps representing the rank performance of \classifier against the two benchmarks for combinations of additive noise and signal length. White implies \classifier outperforms \emph{both} benchmarks, while black implies it is outperformed by at least one of the benchmarks. All 13 system configurations outlined in Table \ref{table:systems} are presented. The majority of each map is white, indicating strong performance of \classifier in a range of situations.}
\label{fig:rank_performance}
\begin{tabular}{ccc}
\includegraphics[scale=0.35]{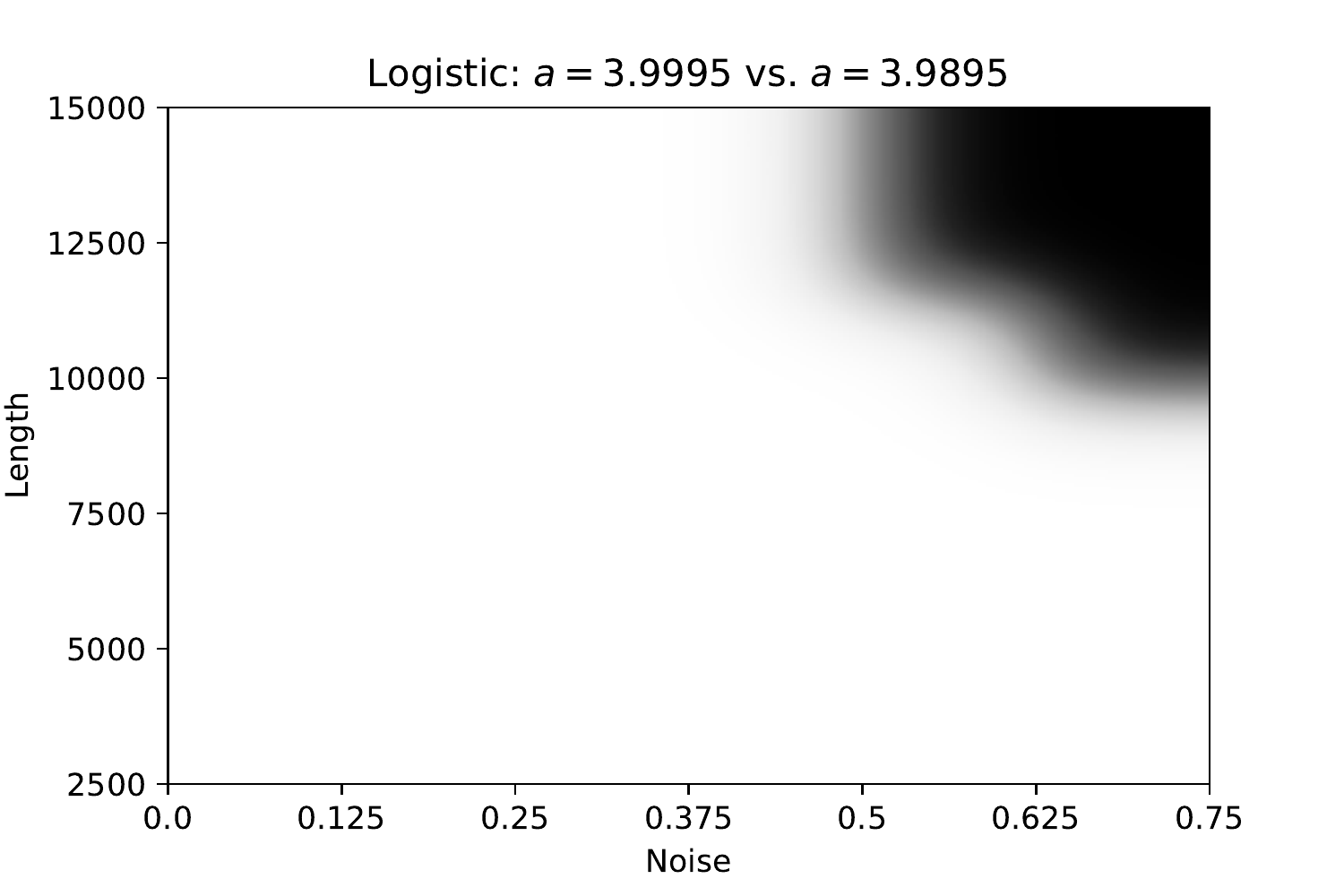} &
\includegraphics[scale=0.35]{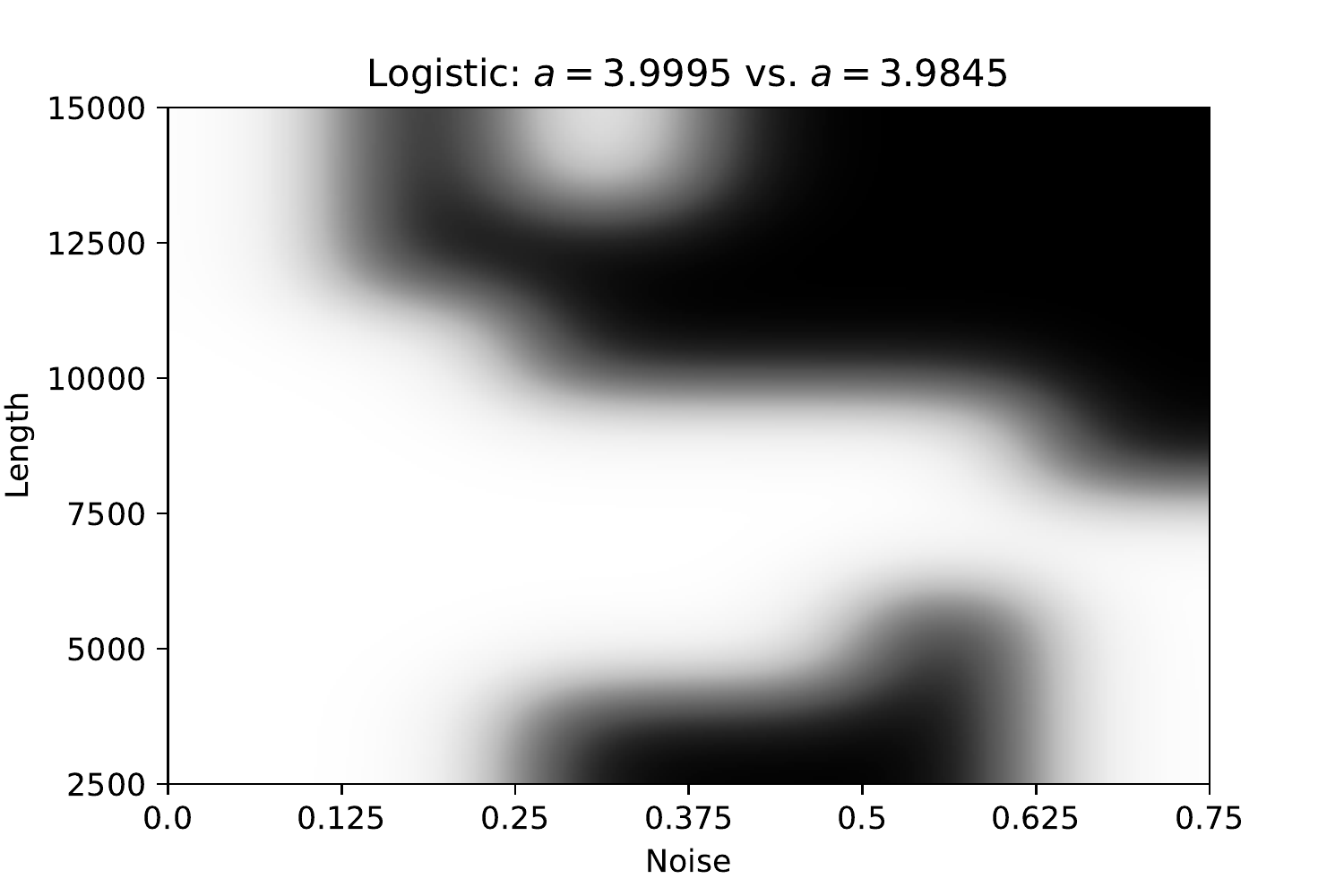} &
\includegraphics[scale=0.35]{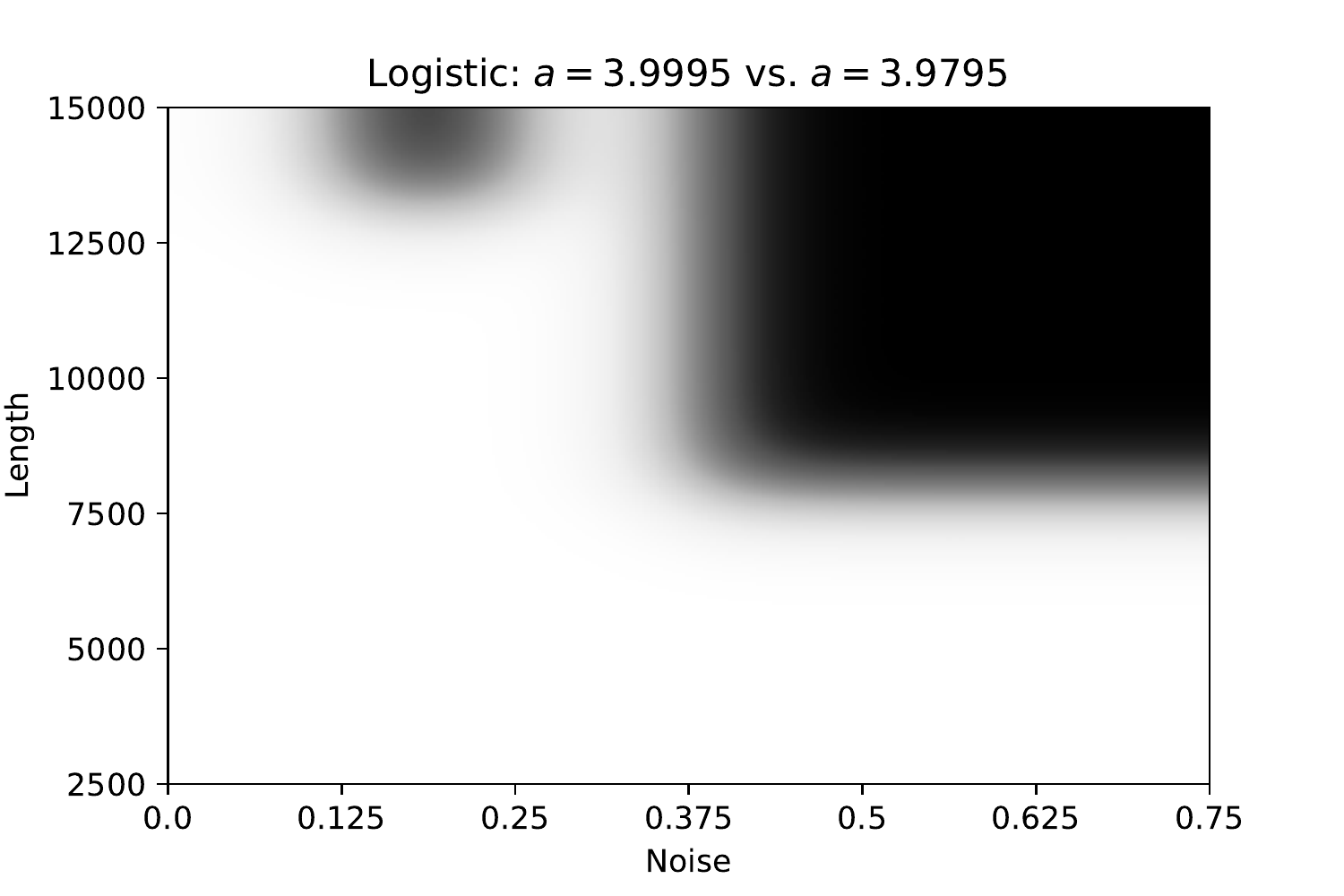} \\
\includegraphics[scale=0.35]{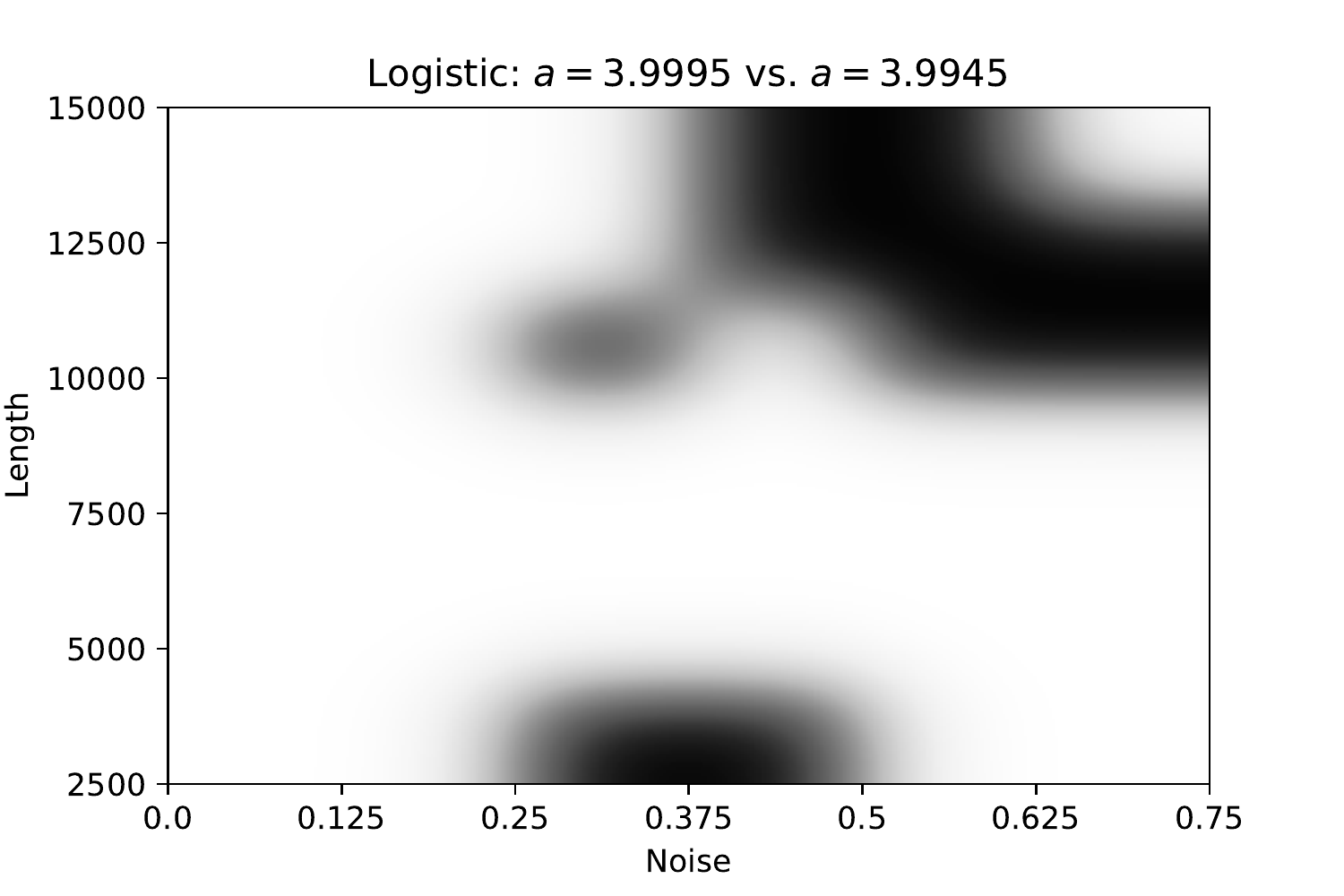} &
\includegraphics[scale=0.35]{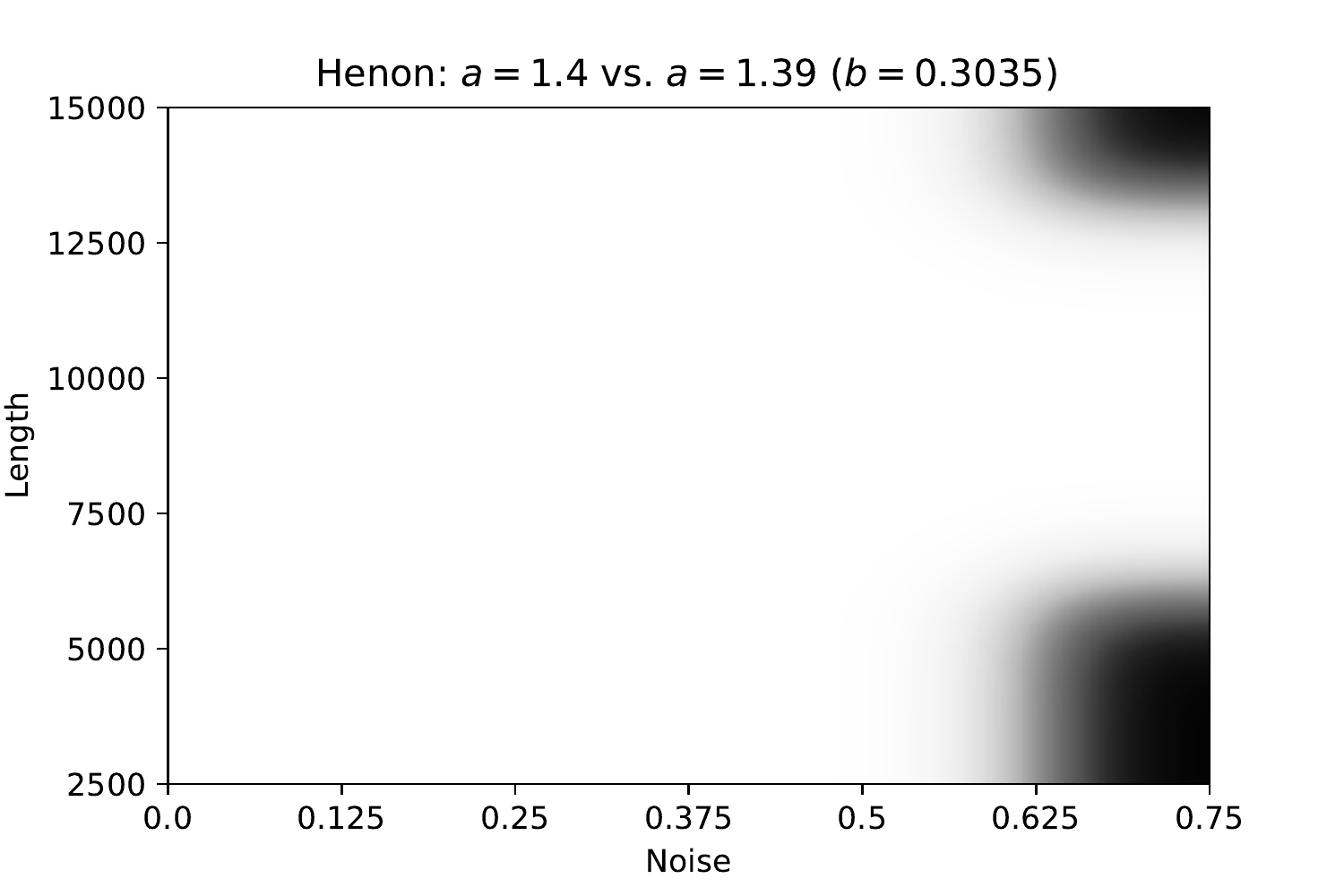} &
\includegraphics[scale=0.35]{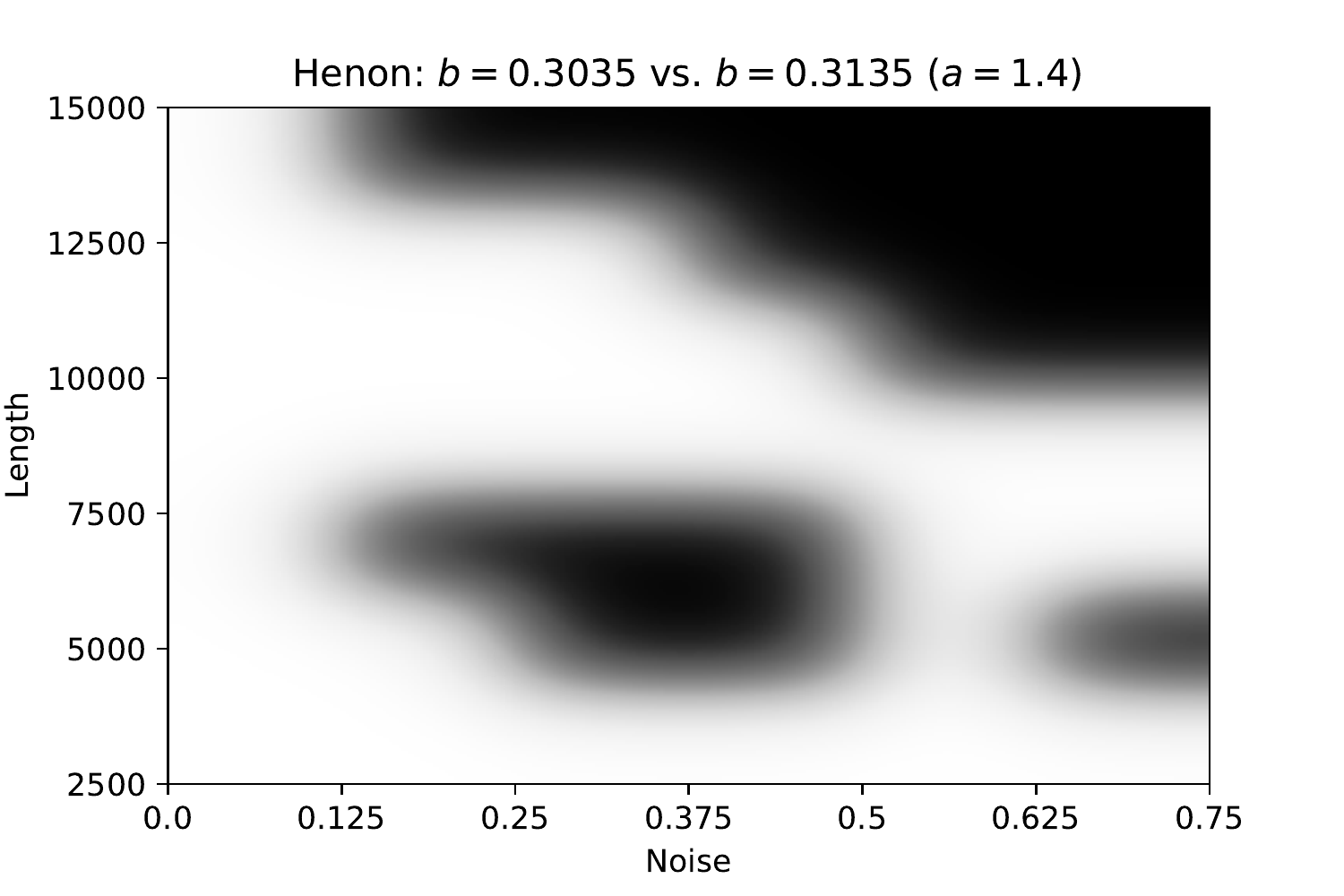} \\
\includegraphics[scale=0.35]{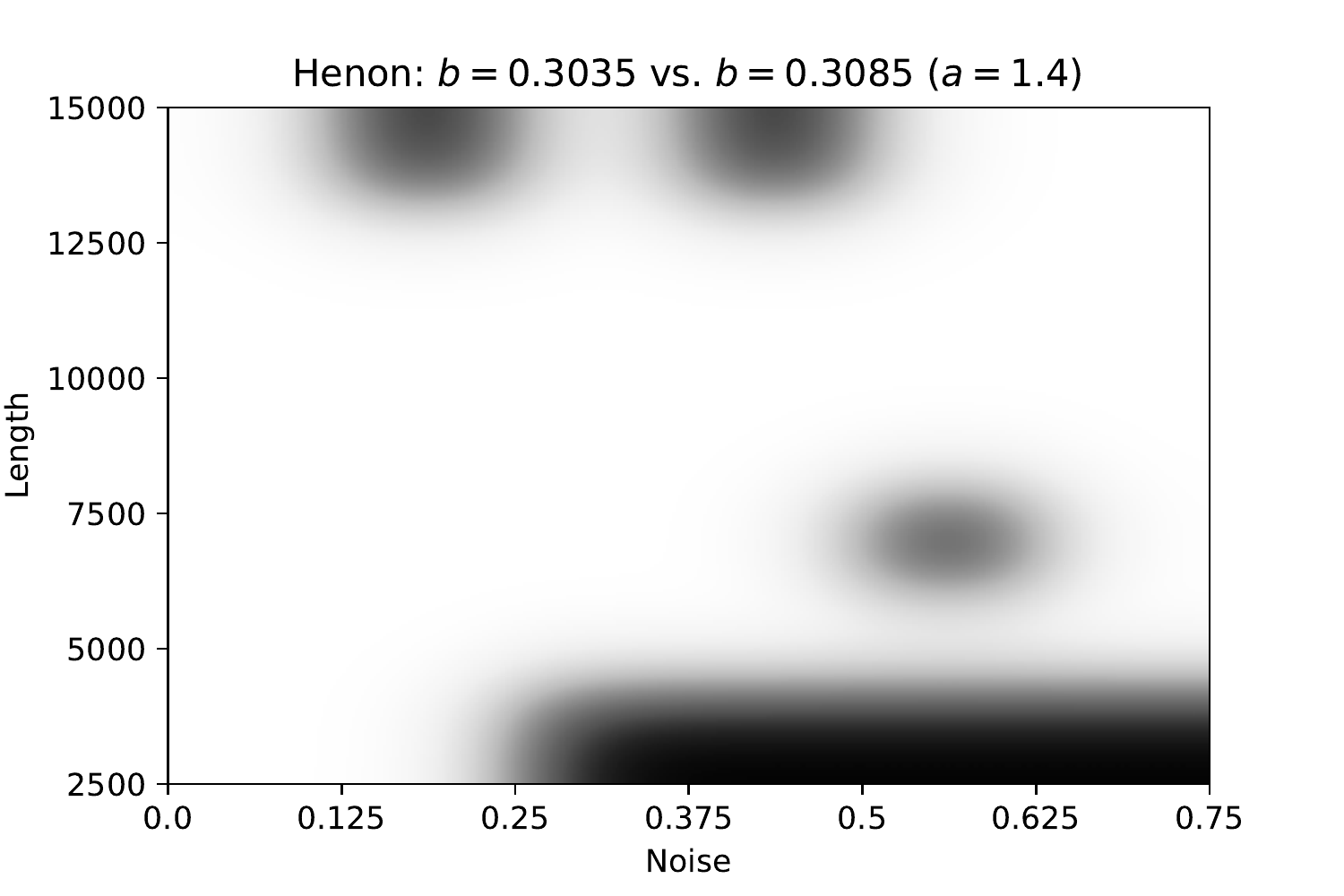} &
\includegraphics[scale=0.35]{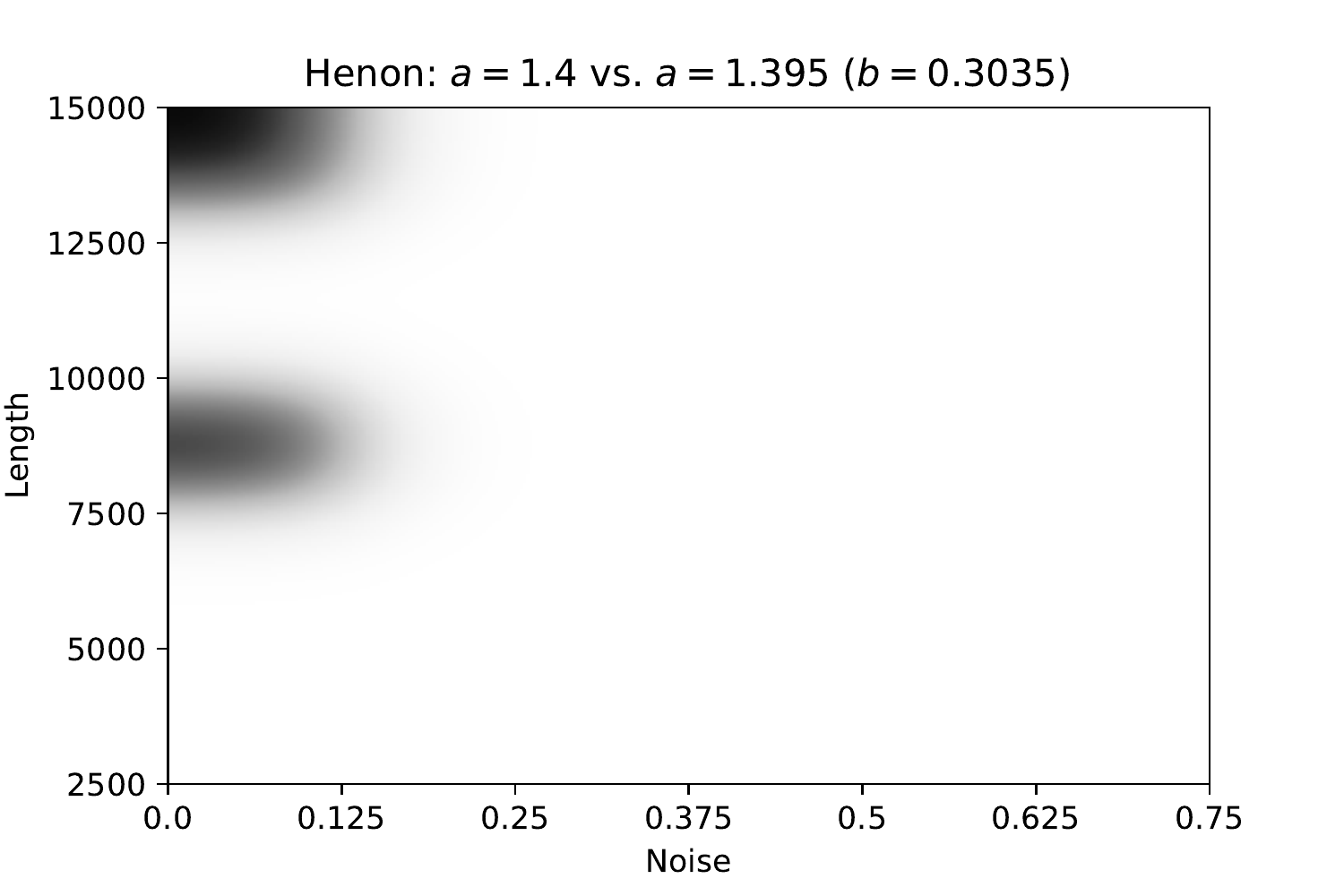} &
\includegraphics[scale=0.35]{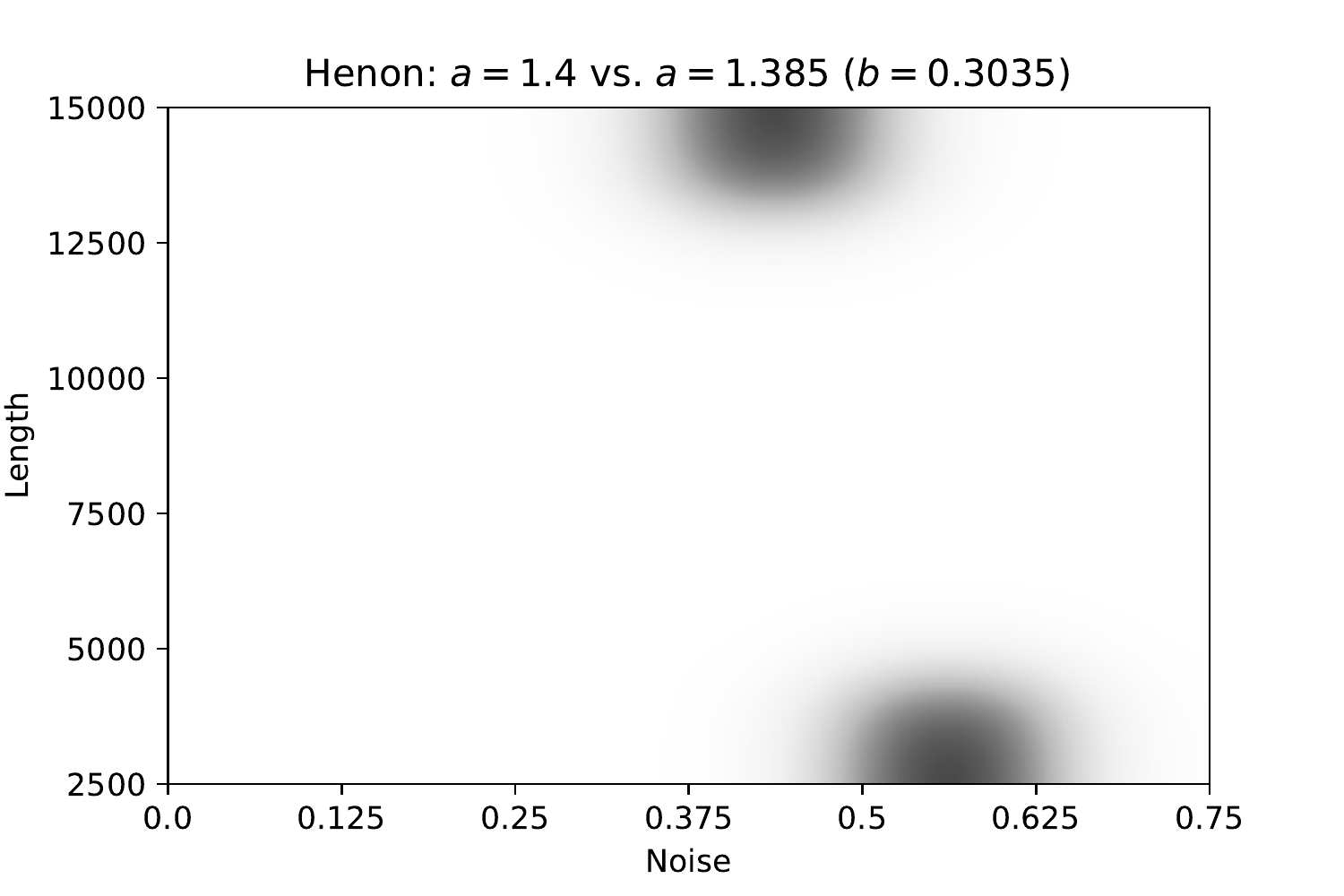} \\
\includegraphics[scale=0.35]{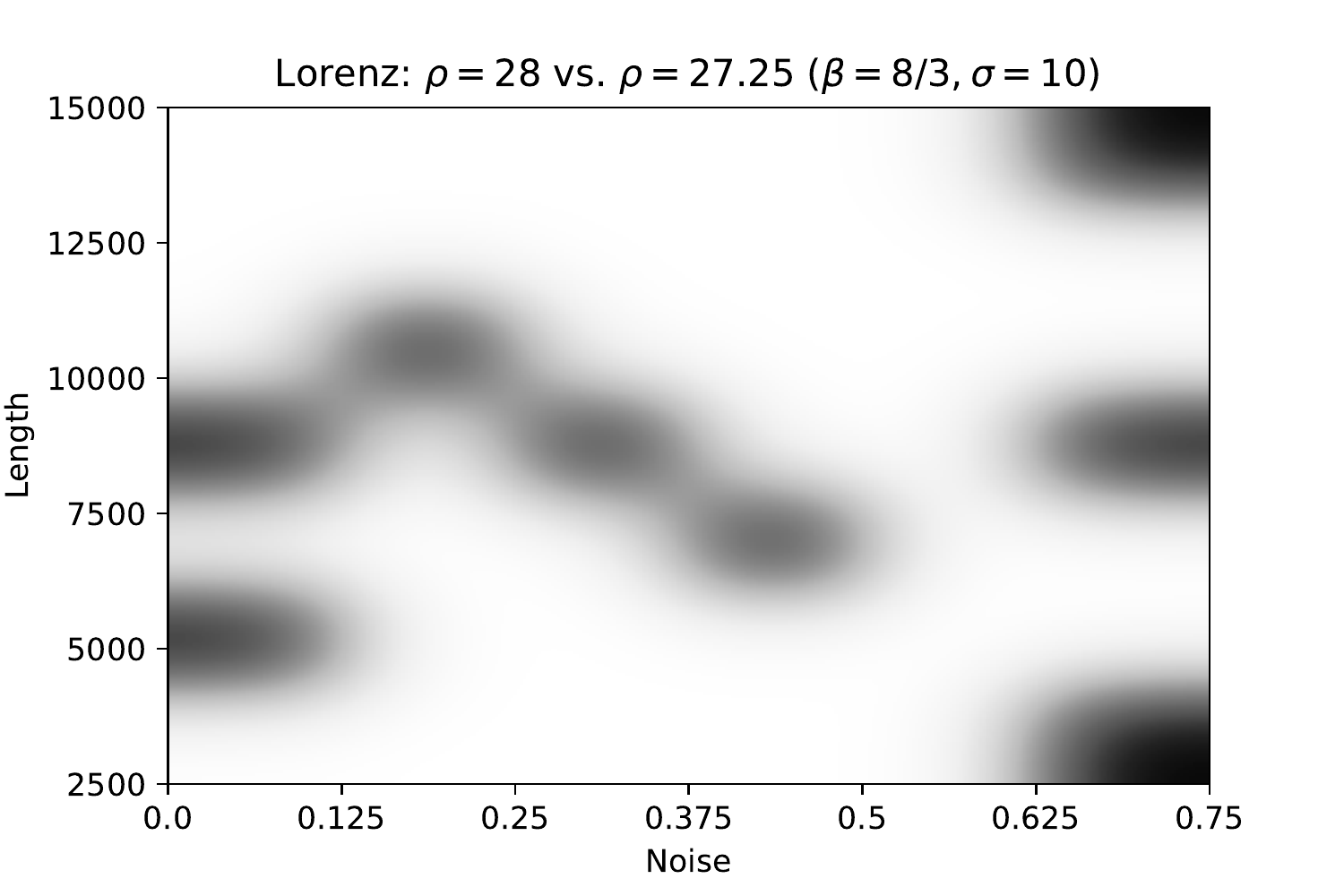} &
\includegraphics[scale=0.35]{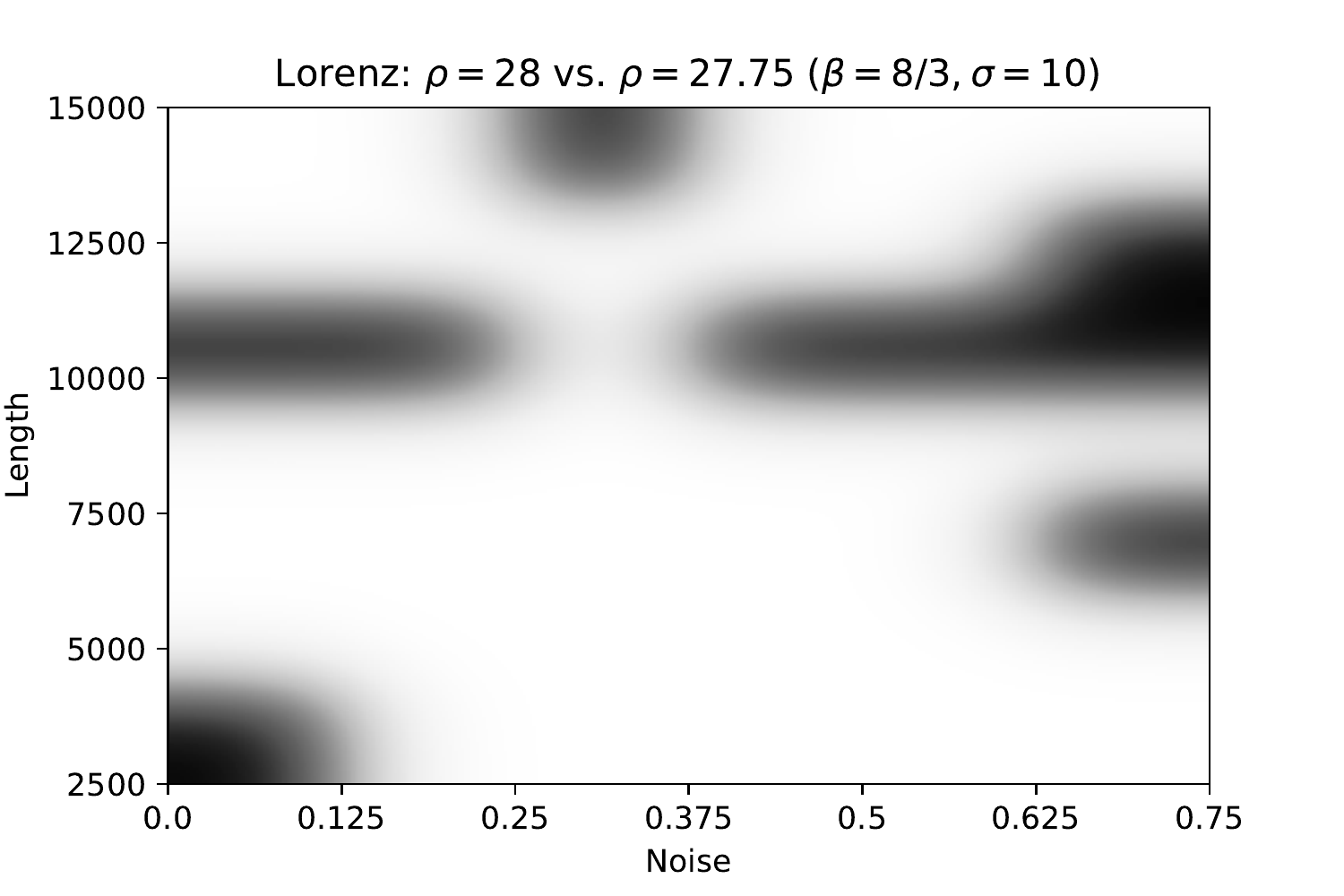} &
\includegraphics[scale=0.35]{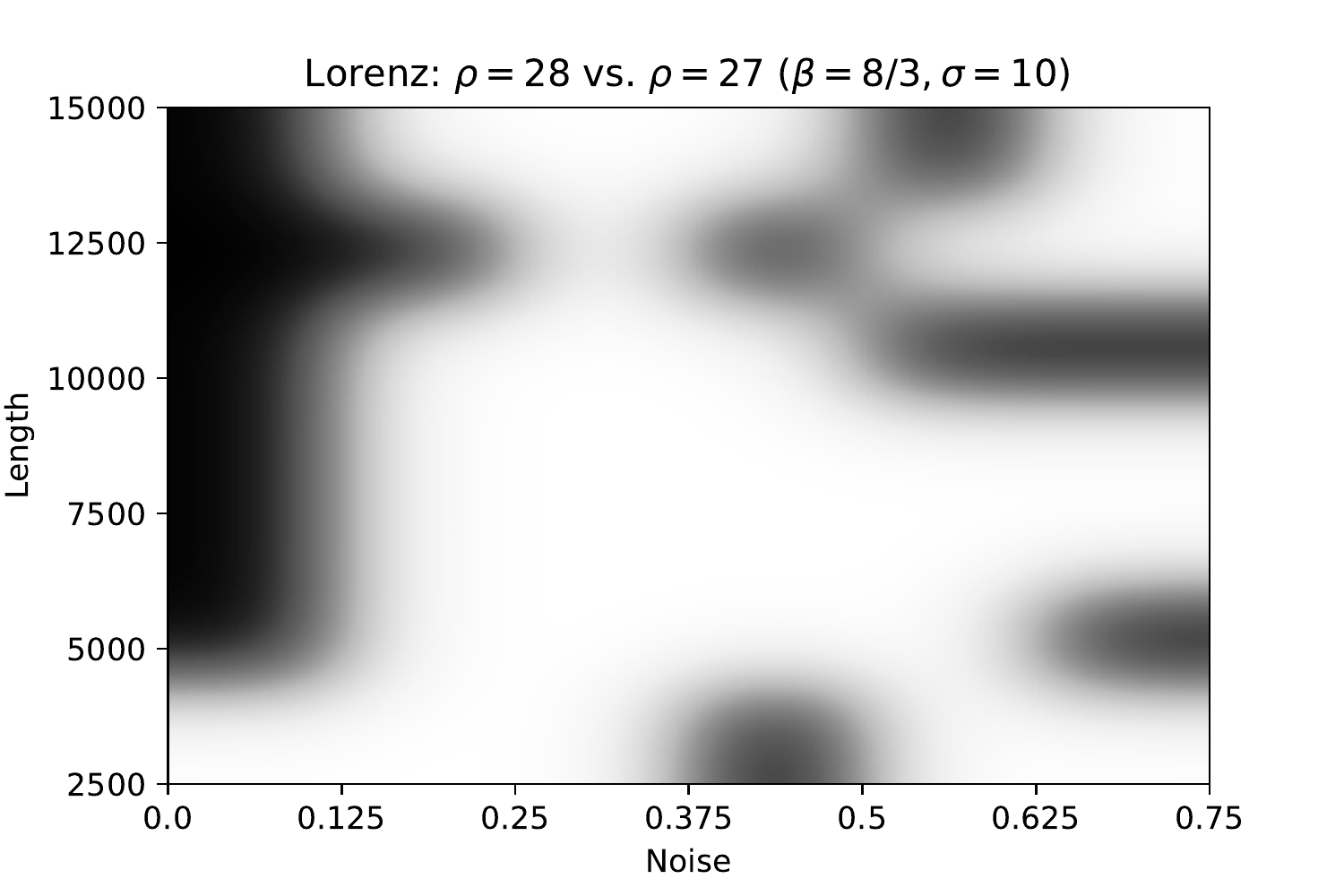} \\
\includegraphics[scale=0.35]{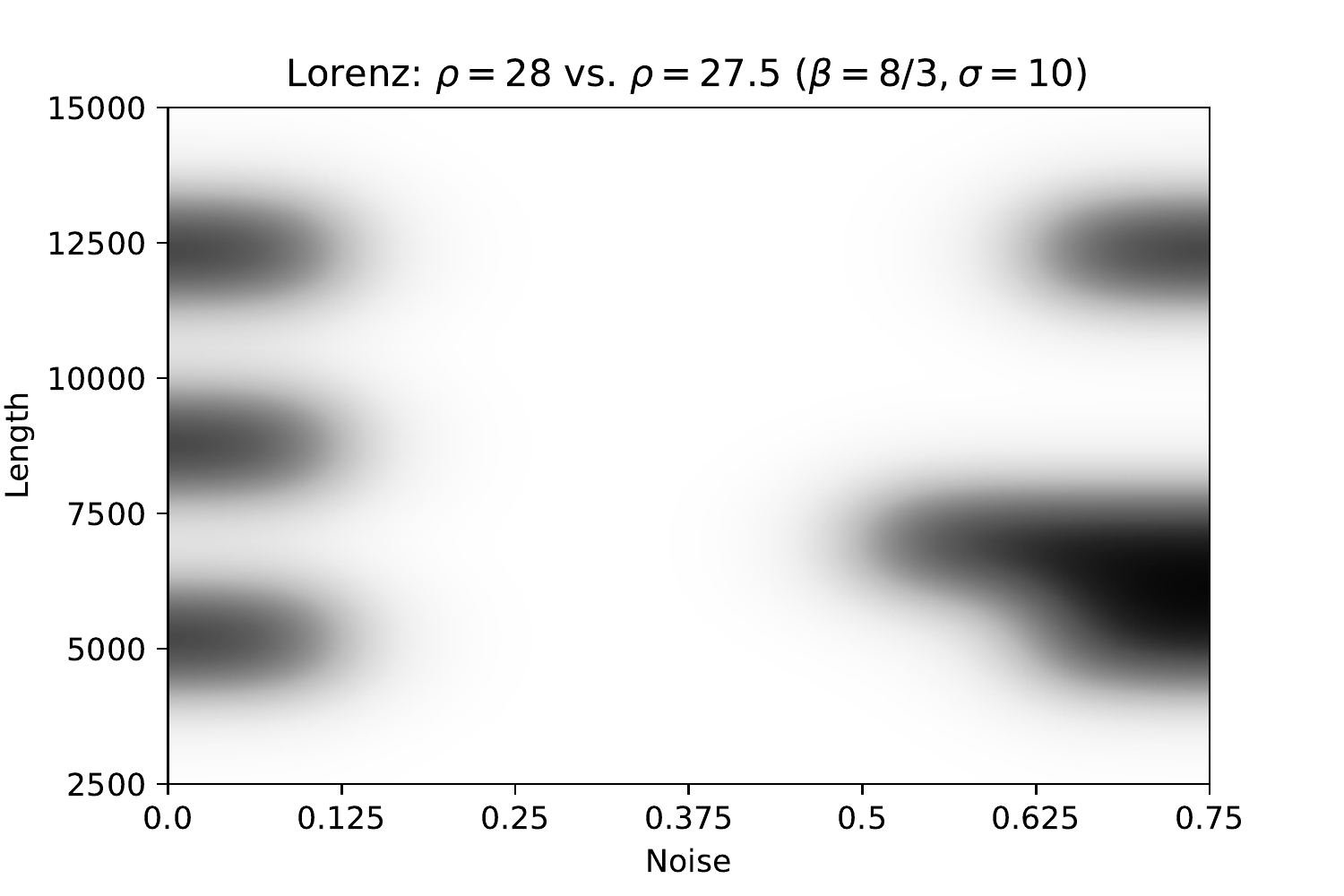} &
\end{tabular}
\end{figure*}

The classifier \classifier outperformed the benchmarks in the vast majority of the experiments, as shown in Figure \ref{fig:rank_performance}. In particular it tended to maintain higher relative accuracy of predictions against the benchmarks as noise levels were increased. This is most easily seen in Figure \ref{fig:accuracy_profiles} where we visualize typical accuracy profiles. In the case of the Lorenz system experiments, \classifier was much more accurate than the PENT classifier, outperformed CEPS on shorter sequences, and matched its performance on longer ones. In experiments generated by the Henon system \classifier outperformed both benchmarks for almost all combinations of sequence length and noise level when the parameter $a$ was varied. This is clearly illustrated in the almost completely white grayscale maps representing the rank performance of \classifier against both of its competitors in Figure \ref{fig:rank_performance}. Another trend visible in Figure \ref{fig:rank_performance} is that the sweet spot for accuracy of \classifier outperforming the others is at mid-range levels of noise (around a standard deviation of 0.375) and mid-range series lengths of around 7500 to 10000 points. Nevertheless white areas constitute over 85\% of these maps on average, suggesting that \classifier is a strong performer in a wide variety of situations.

\begin{figure*}[htb]
\centering
\caption{Short segments from two configurations of a Henon system (top). The associated expected class densities for the full training set (bottom). The latter can help visualize the effects of tuning hyperparameters in \classifier.}
\label{fig:series_snippets}
\hspace{0.5mm}
\includegraphics[width=1\textwidth]{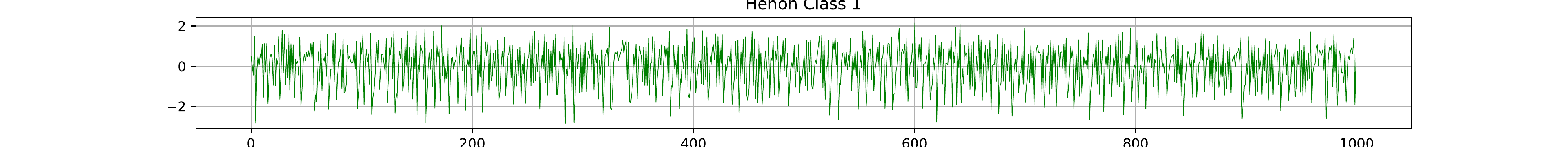} \\
\includegraphics[width=1\textwidth]{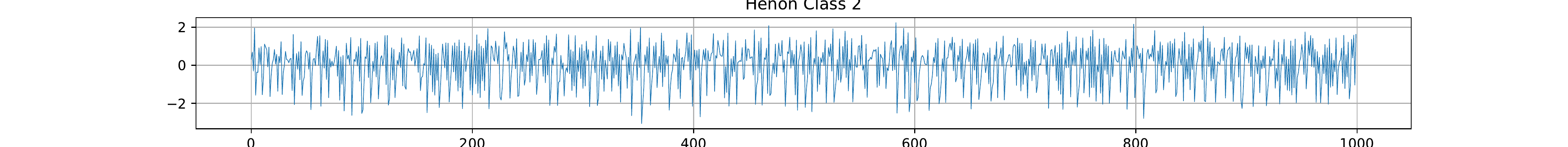} \\
\begin{tabular}{ccc}
\includegraphics[scale=0.35]{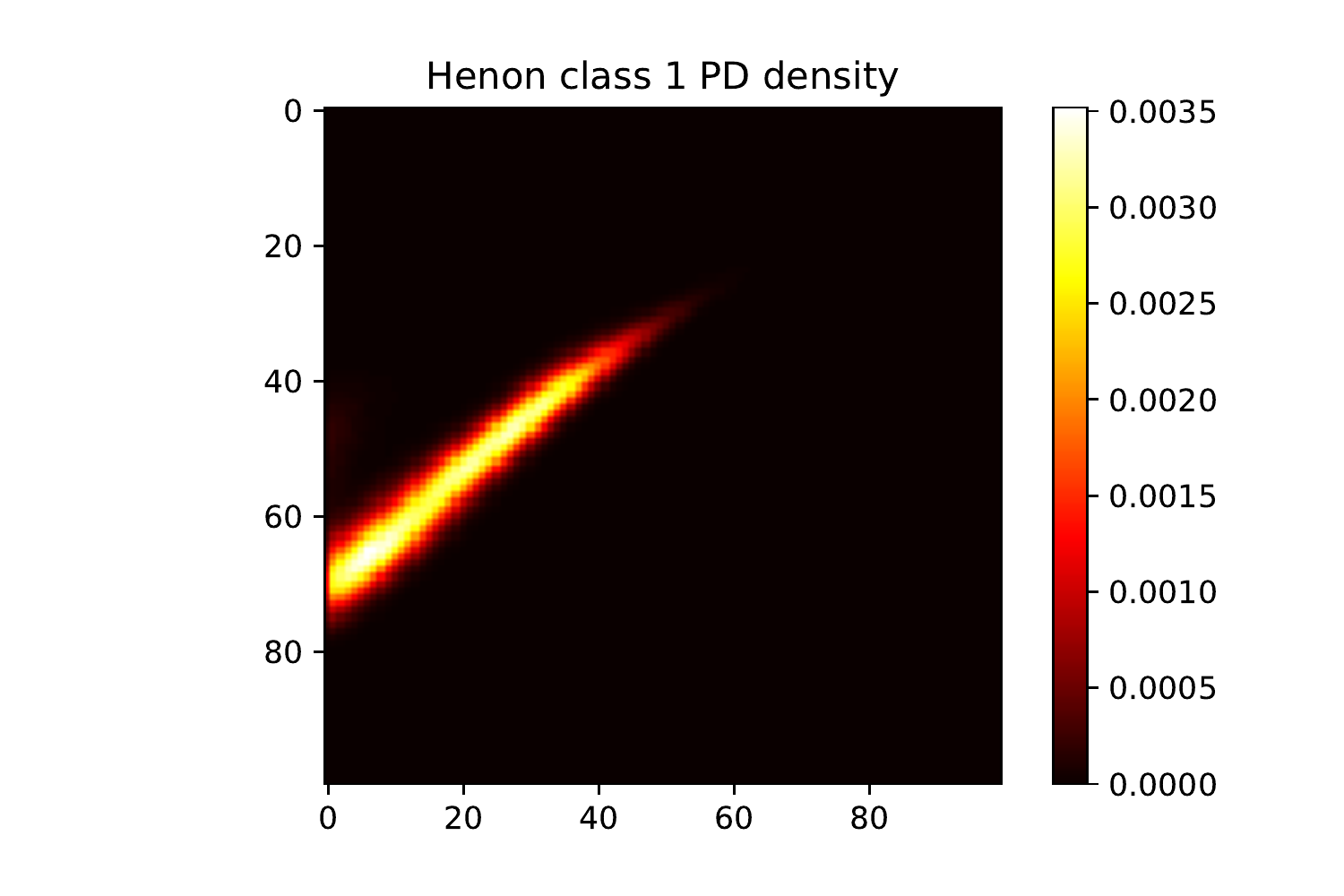} & 
\includegraphics[scale=0.35]{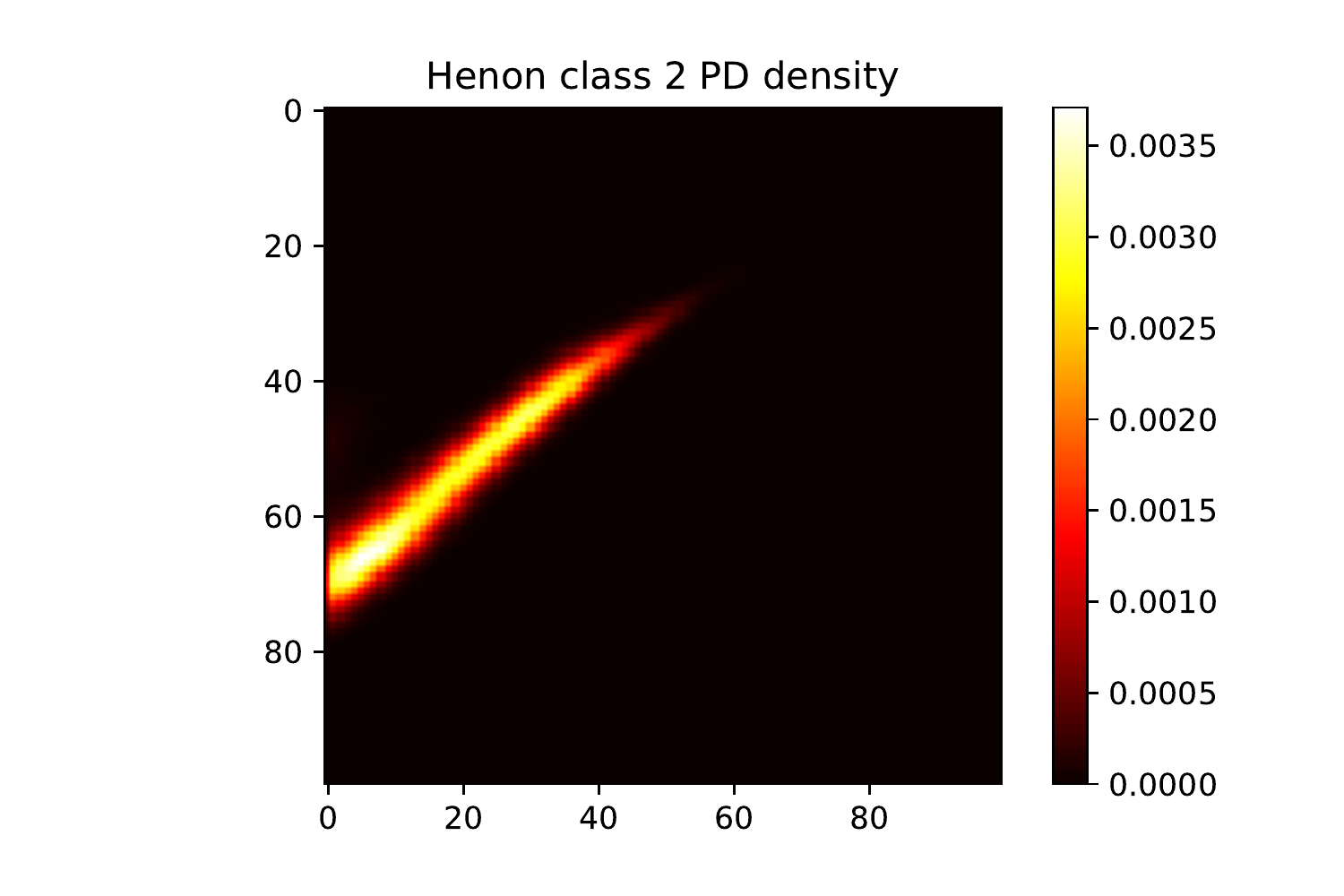} & 
\includegraphics[scale=0.35]{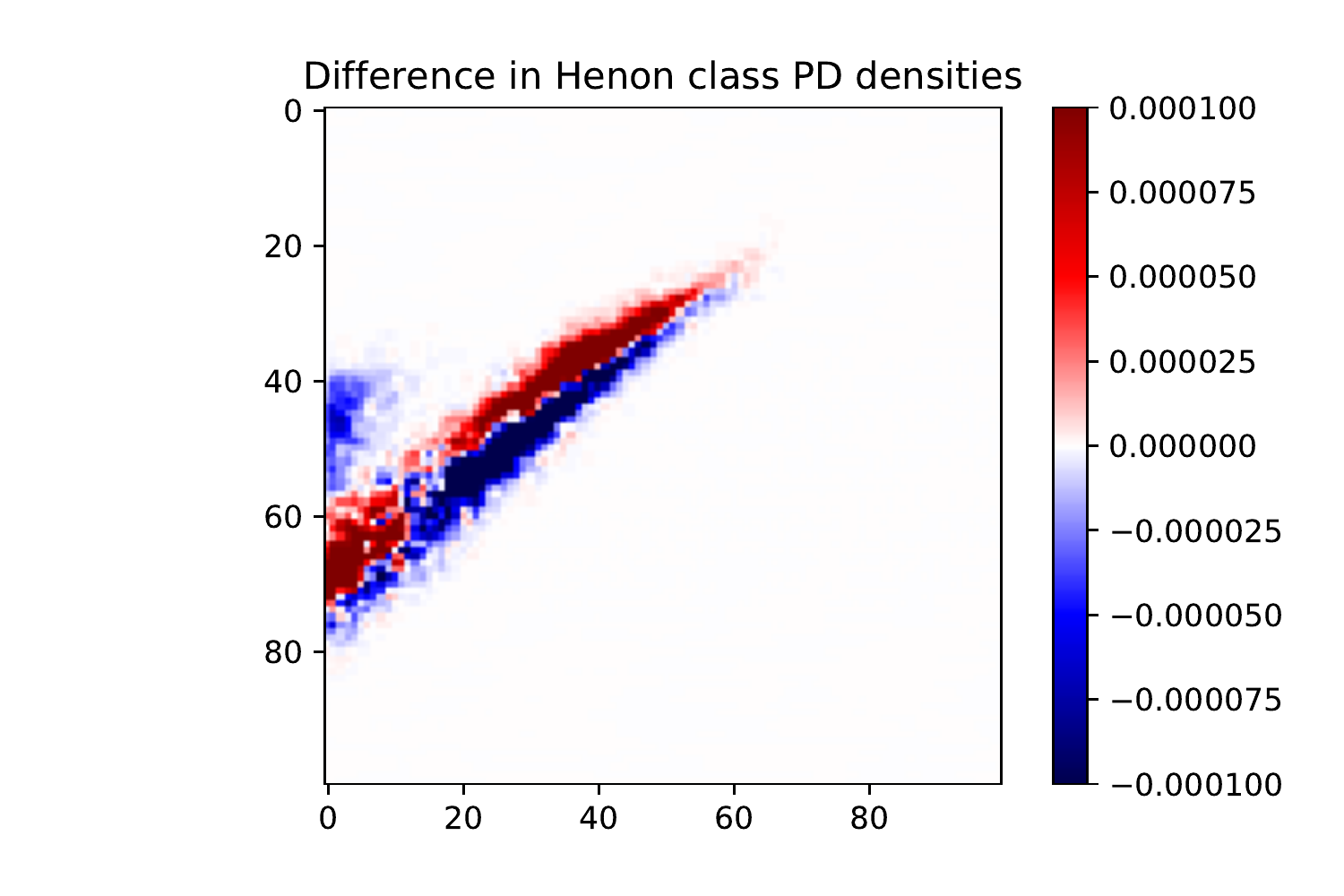} \\ 
\end{tabular}
\end{figure*}

One additional benefit of using \classifier on time series is that it is possible to visualise the topological signatures of the training data classes. This helps to gain an insight into whether or not this topological feature is a `strong candidate' for discriminating classes in a given situation. For example in Figure \ref{fig:series_snippets} we visualise the difference in PD class densities associated to the generated time series snippets. The structure and clear separation of locations of the `plumes' in the third diagram suggest that the estimated PD density is likely to be a good feature in this case, while a more diffuse pattern may indicate that hyperparameters of the model may need to be tuned further or that noise is too dominant.

\section{Conclusion}
\label{sec:conclusion}

We defined a stable and scalable metric on the space of persistence diagrams based on entropic smoothing of optimal transport distances. Following \cite{Adams2017} our approach makes use of weight functions applied to kernel density estimates on PDs to ensure our representations are 1-Wasserstein stable, but we treat the resulting quantized kernel density estimates as two dimensional histograms after renormalization. This allows direct application of regularized transport methods \cite{Cuturi2013a,Solomon2015}.

Unlike existing topological methods for time series \classifier avoids the complexity explosion and noise amplification associated with high dimensional reconstructions based on Takens' embedding theorem. In contrast to traditional signal decomposition methods noise removal is not required for the method to perform well. After cross validation during training the prediction compexity of the model is $O(n\log n)$ with respect to time series length, so it can be applied to long sequences effectively. In addition the underlying metric has a natural parallel implementation (the Sinkhorn Knopp algorithm) suitable for GPUs meaning it is highly scalable. To illustrate these benefits we conducted experiments classifying trajectories of chaotic deterministic systems under a range of signal to noise ratios and parameter values, finding that \classifier is more accurate than two specialized benchmarks in the majority of situations.

\bibliographystyle{IEEEtran}
\bibliography{IEEEabrv,refs}

\begin{thebibliography}{10}
\providecommand{\url}[1]{#1}
\csname url@samestyle\endcsname
\providecommand{\newblock}{\relax}
\providecommand{\bibinfo}[2]{#2}
\providecommand{\BIBentrySTDinterwordspacing}{\spaceskip=0pt\relax}
\providecommand{\BIBentryALTinterwordstretchfactor}{4}
\providecommand{\BIBentryALTinterwordspacing}{\spaceskip=\fontdimen2\font plus
\BIBentryALTinterwordstretchfactor\fontdimen3\font minus
  \fontdimen4\font\relax}
\providecommand{\BIBforeignlanguage}[2]{{%
\expandafter\ifx\csname l@#1\endcsname\relax
\typeout{** WARNING: IEEEtran.bst: No hyphenation pattern has been}%
\typeout{** loaded for the language `#1'. Using the pattern for}%
\typeout{** the default language instead.}%
\else
\language=\csname l@#1\endcsname
\fi
#2}}
\providecommand{\BIBdecl}{\relax}
\BIBdecl

\bibitem{Bagnall2017}
\BIBentryALTinterwordspacing
A.~Bagnall, J.~Lines, A.~Bostrom, J.~Large, and E.~Keogh, ``The great time
  series classification bake off: a review and experimental evaluation of
  recent algorithmic advances,'' \emph{Data Mining and Knowledge Discovery},
  vol.~31, no.~3, pp. 606--660, May 2017. [Online]. Available:
  \url{http://dx.doi.org/10.1007/s10618-016-0483-9}
\BIBentrySTDinterwordspacing

\bibitem{Esling2012}
P.~Esling and C.~Agon, ``{Time-series data mining},'' \emph{ACM Computing
  Surveys (CSUR)}, vol.~45, no.~1, pp. 1--34, 2012.

\bibitem{Kantz2004}
H.~Kantz and T.~Schreiber, \emph{Nonlinear time series analysis}.\hskip 1em
  plus 0.5em minus 0.4em\relax Cambridge university press, 2004, vol.~7.

\bibitem{Koopmans1995}
L.~H. Koopmans, \emph{The spectral analysis of time series}.\hskip 1em plus
  0.5em minus 0.4em\relax Elsevier, 1995.

\bibitem{Randall2017}
R.~B. Randall, ``A history of cepstrum analysis and its application to
  mechanical problems,'' \emph{Mechanical Systems and Signal Processing},
  vol.~97, pp. 3--19, 2017.

\bibitem{Carlsson2009}
\BIBentryALTinterwordspacing
G.~Carlsson, ``Topology and data,'' \emph{Bulletin of the American Mathematical
  Society}, pp. 1--49, 2009. [Online]. Available:
  \url{http://www.ams.org/journals/bull/2009-46-02/S0273-0979-09-01249-X/}
\BIBentrySTDinterwordspacing

\bibitem{Edelsbrunner2014}
H.~Edelsbrunner and D.~Morozov, ``{Persistent Homology : Theory and
  Practice},'' 2014.

\bibitem{Cohen-Steiner2010}
D.~Cohen-Steiner, H.~Edelsbrunner, J.~Harer, and Y.~Mileyko, ``{Lipschitz
  Functions Have $L_p$-Stable Persistence},'' \emph{Foundations of
  Computational Mathematics}, vol.~10, pp. 127--139, 2010.

\bibitem{Ghrist2008a}
\BIBentryALTinterwordspacing
R.~Ghrist, ``{Barcodes: the persistent topology of data},'' \emph{Bulletin of
  the American Mathematical Society}, vol.~45, no.~1, pp. 61--75, 2008.
  [Online]. Available:
  \url{http://www.ams.org/bull/2008-45-01/S0273-0979-07-01191-3/}
\BIBentrySTDinterwordspacing

\bibitem{Edelsbrunner2010}
H.~Edelsbrunner and J.~Harer, \emph{Computational topology: An
  introduction}.\hskip 1em plus 0.5em minus 0.4em\relax American Mathematical
  Society, 2010.

\bibitem{Adams2017}
H.~Adams, T.~Emerson, M.~Kirby, R.~Neville, C.~Peterson, P.~Shipman,
  S.~Chepushtanova, E.~Hanson, F.~Motta, and L.~Ziegelmeier, ``Persistence
  images: A stable vector representation of persistent homology,'' \emph{The
  Journal of Machine Learning Research}, vol.~18, no.~1, pp. 218--252, 2017.

\bibitem{Reininghaus2015}
\BIBentryALTinterwordspacing
J.~Reininghaus, S.~Huber, U.~Bauer, and R.~Kwitt, ``{A Stable Multi-Scale
  Kernel for Topological Machine Learning},'' in \emph{Conference on Computer
  Vision and Pattern Recognition (CVPR 2015)}.\hskip 1em plus 0.5em minus
  0.4em\relax IEEE, 2015, pp. 4741--4748. [Online]. Available:
  \url{http://www.cv-foundation.org/openaccess/content\_cvpr\_2015/papers/Reininghaus\_A\_Stable\_Multi-Scale\_2015\_CVPR\_paper.pdf}
\BIBentrySTDinterwordspacing

\bibitem{Khasawneh2016}
\BIBentryALTinterwordspacing
F.~A. Khasawneh and E.~Munch, ``{Chatter detection in turning using persistent
  homology},'' \emph{Mechanical Systems and Signal Processing}, vol. 70-71, pp.
  527--541, 2016. [Online]. Available:
  \url{http://www.sciencedirect.com/science/article/pii/S0888327015004598}
\BIBentrySTDinterwordspacing

\bibitem{Perea2014}
J.~A. Perea and J.~Harer, ``{Sliding Windows and Persistence: An Application of
  Topological Methods to Signal Analysis},'' \emph{Foundations of Computational
  Mathematics}, vol.~15, no.~3, pp. 799--838, 2014.

\bibitem{Pereira2015a}
C.~M.~M. Pereira and R.~F. de~Mello, ``{Persistent homology for time series and
  spatial data clustering},'' \emph{Expert Systems with Applications}, vol.~42,
  pp. 6026--6038, 2015.

\bibitem{Seversky2016}
L.~M. Seversky, S.~Davis, and M.~Berger, ``{On Time-series Topological Data
  Analysis: New Data and Opportunities},'' \emph{The IEEE Conference on
  Computer Vision and Pattern Recognition (CVPR) Workshops}, pp. 59--67, 2016.

\bibitem{Alexander2015}
\BIBentryALTinterwordspacing
Z.~Alexander, E.~Bradley, J.~D. Meiss, and N.~Sanderson, ``Simplicial
  multivalued maps and the witness complex for dynamical analysis of time
  series,'' \emph{SIAM J. Applied Dynamical Systems}, vol.~14, no.~3, pp.
  1278--1307, 2015. [Online]. Available: \url{http://arxiv.org/abs/1406.2245}
\BIBentrySTDinterwordspacing

\bibitem{DeSilva2004}
\BIBentryALTinterwordspacing
V.~De~Silva and G.~Carlsson, ``Topological estimation using witness
  complexes,'' in \emph{Proceedings of the First Eurographics Conference on
  Point-Based Graphics}, ser. SPBG'04.\hskip 1em plus 0.5em minus 0.4em\relax
  Aire-la-Ville, Switzerland, Switzerland: Eurographics Association, 2004, pp.
  157--166. [Online]. Available:
  \url{http://dx.doi.org/10.2312/SPBG/SPBG04/157-166}
\BIBentrySTDinterwordspacing

\bibitem{Sanderson2017}
N.~{Sanderson}, E.~{Shugerman}, S.~{Molnar}, J.~D. {Meiss}, and E.~{Bradley},
  ``{Computational Topology Techniques for Characterizing Time-Series Data},''
  \emph{ArXiv e-prints}, Aug. 2017.

\bibitem{Rucco2017}
M.~Rucco, R.~Gonzalez-Diaz, M.-J. Jimenez, N.~Atienza, C.~Cristalli,
  E.~Concettoni, A.~Ferrante, and E.~Merelli, ``A new topological entropy-based
  approach for measuring similarities among piecewise linear functions,''
  \emph{Signal Processing}, vol. 134, pp. 130--138, 2017.

\bibitem{Chazal2018}
F.~{Chazal} and V.~{Divol}, ``{The density of expected persistence diagrams and
  its kernel based estimation},'' \emph{ArXiv e-prints}, Feb. 2018.

\bibitem{Cuturi2013a}
M.~Cuturi, ``Sinkhorn distances: Lightspeed computation of optimal transport,''
  in \emph{Advances in neural information processing systems}, 2013, pp.
  2292--2300.

\bibitem{Altschuler2017}
\BIBentryALTinterwordspacing
J.~Altschuler, J.~Weed, and P.~Rigollet, ``Near-linear time approximation
  algorithms for optimal transport via {S}inkhorn iteration,'' \emph{CoRR},
  vol. abs/1705.09634, 2017. [Online]. Available:
  \url{http://arxiv.org/abs/1705.09634}
\BIBentrySTDinterwordspacing

\bibitem{Blondel2017}
M.~{Blondel}, V.~{Seguy}, and A.~{Rolet}, ``{Smooth and Sparse Optimal
  Transport},'' \emph{ArXiv e-prints}, Oct. 2017.

\bibitem{Solomon2015}
J.~Solomon, F.~De~Goes, G.~Peyr{\'e}, M.~Cuturi, A.~Butscher, A.~Nguyen, T.~Du,
  and L.~Guibas, ``Convolutional {W}asserstein distances: Efficient optimal
  transportation on geometric domains,'' \emph{ACM Transactions on Graphics
  (TOG)}, vol.~34, no.~4, p.~66, 2015.

\bibitem{Lacombe2018}
T.~{Lacombe}, M.~{Cuturi}, and S.~{Oudot}, ``{Large Scale computation of Means
  and Clusters for Persistence Diagrams using Optimal Transport},'' \emph{ArXiv
  e-prints}, May 2018.

\bibitem{Marwan2007}
N.~Marwan, M.~C. Romano, M.~Thiel, and J.~Kurths, ``Recurrence plots for the
  analysis of complex systems,'' \emph{Physics reports}, vol. 438, no. 5-6, pp.
  237--329, 2007.

\bibitem{Wolf1985}
A.~Wolf, J.~B. Swift, H.~L. Swinney, and J.~A. Vastano, ``Determining lyapunov
  exponents from a time series,'' \emph{Physica D: Nonlinear Phenomena},
  vol.~16, no.~3, pp. 285--317, 1985.

\bibitem{Cohen-Steiner2007}
\BIBentryALTinterwordspacing
D.~Cohen-Steiner, H.~Edelsbrunner, and J.~Harer, ``Stability of persistence
  diagrams,'' \emph{Discrete {\&} Computational Geometry}, vol.~37, no.~1, pp.
  103--120, 2007. [Online]. Available:
  \url{http://dx.doi.org/10.1007/s00454-006-1276-5}
\BIBentrySTDinterwordspacing

\bibitem{Edelsbrunner2008}
H.~Edelsbrunner and J.~Harer, ``{Persistent homology---a survey},'' in
  \emph{Surveys on discrete and computational geometry: twenty years later:
  AMS-IMS-SIAM Joint Summer Research Conference, June 18-22, 2006, Snowbird,
  Utah}, vol. 453, 2008, pp. 257--282.

\bibitem{dAmico2003}
M.~d'Amico, P.~Frosini, and C.~Landi, ``Optimal matching between reduced size
  functions,'' \emph{DISMI, Univ. di Modena e Reggio Emilia, Italy, Technical
  report}, no.~35, 2003.

\bibitem{Vejdemo-Johansson2012}
\BIBentryALTinterwordspacing
M.~Vejdemo-Johansson, ``{Sketches of a platypus: persistent homology and its
  algebraic foundations},'' \emph{arXiv preprint arXiv:1212.5398}, pp. 1--22,
  2012. [Online]. Available: \url{http://arxiv.org/abs/1212.5398}
\BIBentrySTDinterwordspacing

\bibitem{Kerber2016a}
M.~Kerber, D.~Morozov, and A.~Nigmetov, ``Geometry helps to compare persistence
  diagrams,'' \emph{ArXiv e-prints}, 2016.

\bibitem{Turner2013}
\BIBentryALTinterwordspacing
K.~Turner, ``{Means and medians of sets of persistence diagrams},'' 2013.
  [Online]. Available: \url{http://arxiv.org/abs/1307.8300}
\BIBentrySTDinterwordspacing

\bibitem{Villani2003}
C.~Villani, \emph{Topics in optimal transportation}.\hskip 1em plus 0.5em minus
  0.4em\relax American Mathematical Society, 2003.

\bibitem{Villani2008}
------, \emph{Optimal transport: old and new}.\hskip 1em plus 0.5em minus
  0.4em\relax Springer Science \& Business Media, 2008, vol. 338.

\bibitem{Avis1980}
D.~Avis, ``On the extreme rays of the metric cone,'' \emph{Canad. J. Math},
  vol.~32, no.~1, pp. 126--144, 1980.

\bibitem{Pele2009}
O.~Pele and M.~Werman, ``Fast and robust earth mover's distances,'' in
  \emph{Computer vision, 2009 IEEE 12th international conference on}.\hskip 1em
  plus 0.5em minus 0.4em\relax IEEE, 2009, pp. 460--467.

\bibitem{Benamou2015}
J.-D. Benamou, G.~Carlier, M.~Cuturi, L.~Nenna, and G.~Peyr{\'e}, ``Iterative
  bregman projections for regularized transportation problems,'' \emph{SIAM
  Journal on Scientific Computing}, vol.~37, no.~2, pp. A1111--A1138, 2015.

\bibitem{Dessein2016}
A.~Dessein, N.~Papadakis, and J.-L. Rouas, ``Regularized optimal transport and
  the {ROT} mover's distance,'' \emph{arXiv preprint arXiv:1610.06447}, 2016.

\bibitem{Genevay2017}
A.~Genevay, G.~Peyr{\'e}, and M.~Cuturi, ``Learning generative models with
  sinkhorn divergences,'' \emph{arXiv preprint arXiv:1706.00292}, 2017.

\bibitem{Vogel2002}
C.~R. Vogel, \emph{Computational methods for inverse problems}.\hskip 1em plus
  0.5em minus 0.4em\relax Siam, 2002, vol.~23.

\bibitem{Rucco2015}
M.~Rucco, E.~Concettoni, C.~Cristalli, A.~Ferrante, and E.~Merelli,
  ``{Topological classification of small DC motors},'' \emph{2015 IEEE 1st
  International Forum on Research and Technologies for Society and Industry,
  RTSI 2015 - Proceedings}, pp. 192--197, 2015.

\bibitem{Kalpakis2001}
K.~Kalpakis, D.~Gada, and V.~Puttagunta, ``Distance measures for effective
  clustering of arima time-series,'' in \emph{Proceedings 2001 IEEE
  International Conference on Data Mining}, 2001, pp. 273--280.

\end{thebibliography}

\end{document}